\documentclass[10pt,twocolumn,letterpaper]{article}

\usepackage{cvpr}
\usepackage{times}
\usepackage{url}
\usepackage{epsfig}
\usepackage{graphicx}
\usepackage{amsmath}
\usepackage{amssymb}
\usepackage{float}
\usepackage{epstopdf}
\usepackage{multirow}
\usepackage{subcaption}

\graphicspath{{figures/}}

\cvprfinalcopy 

\def\cvprPaperID{****} 

\ifcvprfinal\fi
\begin{document}

\title{LOGO-Net: Large-scale Deep Logo Detection and Brand Recognition\\with Deep Region-based Convolutional Networks}

\author{Steven C.H. Hoi$^\star$, Xiongwei Wu$^\star$, Hantang Liu$^\star$, Yue Wu$^\star$, Huiqiong Wang$^\ddagger$, Hui Xue$^\ddagger$, Qiang Wu$^\ddagger$\\
$^\star$School of Information Systems, Singapore Management University, Singapore\\
$^\ddagger$Alibaba Group, Hangzhou, China\\
{\tt\small \{chhoi,yuewu,htliu,xwwu.2015@phdis\}@smu.edu.sg}\\
{\tt\small\{huiqiong.whq,hui.xueh,qiangwu.wq\}@alibaba-inc.com}\\
}

\maketitle

\begin{abstract}
Logo detection from images has many applications, particularly for brand recognition and intellectual property protection. Most existing studies for logo recognition and detection are based on small-scale datasets which are not comprehensive enough when exploring emerging deep learning techniques. In this paper, we introduce ``LOGO-Net"\footnote{The LOGO-net will be released at \url{http://logo-net.org/}\\
This project was initialized in early of 2015, and the main tasks were completed in July 2015 when Prof Hoi visited Alibaba Group.}, a large-scale logo image database for logo detection and brand recognition from real-world product images. To facilitate research, LOGO-Net has two datasets: (i)``logos-18" consists of 18 logo classes, 10 brands, and 16,043 logo objects, and (ii) ``logos-160" consists of 160 logo classes, 100 brands, and 130,608 logo objects. We describe the ideas and challenges for constructing such a large-scale database. Another key contribution of this work is to apply emerging deep learning techniques for logo detection and brand recognition tasks, and conduct extensive experiments by exploring several state-of-the-art deep region-based convolutional networks techniques for object detection tasks.
\end{abstract}

\vspace{-0.2in}
\section{Introduction}\label{section: introduction}


Logo detection and recognition has been extensively studied in computer vision and pattern recognition literature~\cite{doermann1993logo,cesarini1997neural,francesconi1998logo,chen2003noisy,den2003logo,zhu2007automatic,kleban2008spatial,belgalogos2009,gao2009logo,li2010fast,psyllos2010vehicle,RombergICMR2011,logo-icmr2011}. From a computer vision perspective, {\it logo recognition}, which can be viewed as a special case of image recognition, aims to recognize the logo name of an input image, and {\it logo detection} is often more challenging in that it not only needs to recognize the logo name but also need to find the locations of logo objects in the input image. Logo detection and recognition found a wide range of applications in many domains, such as product brand recognition for intellectual property protection in e-commerce platforms, vehicle logo recognition for intelligent transportation~\cite{psyllos2010vehicle}, product brand management on social media~\cite{gao2014brand}, etc.

Logo objects typically consist of mixed text and graphic symbols. Although it may be viewed as a special type of object detection, logo detection from real-world images (e.g., product images) can be quite challenging since the same logo when appearing in different real scenarios can become very different due to the changes in sizes, rotations, lighting, occlusion, rigid and even non-rigid transformations. For example, a rigid logo object when appearing in a real clothing image often becomes non-rigid, making it difficult to be detected and recognized.

Although logo-related research has been explored for a long history of over two decades in literature~\cite{doermann1993logo,cesarini1997neural,francesconi1998logo}, most existing studies use small datasets and very few large datasets are publicly available. For example, among the existing publicly available logo databases~\cite{belgalogos2009}, one of the largest is ``FlickrLogos-32", which only consists of 32 logo classes, 8240 images, and 5644 logo objects. Clearly, the existing datasets are not sufficient for conducting large-scale logo related research, particularly when exploring emerging data-intensive deep learning techniques.

To this end, we propose ``LOGO-Net" --- a large-scale logo image database to facilitate the research of logo detection and product brand recognition. The current LOGO-Net database consists of 160 logo classes, 100 brands, 73,414 images, and a total of 130,608 logo objects manually labeled with bounding boxes by human beings. Constructing such a large-scale database is challenging, time-consuming, and expensive. We discuss the details of how to construct the database and resolve the challenges in the project. In addition to the database, another key contribution of this work is to explore a family of emerging state-of-the-art deep learning techniques for generic object detection with application to large-scale logo detection and brand recognition tasks and conduct extensive empirical evaluations.

\begin{figure*}[htb]
\vspace{-0.1in}
\begin{center}
\includegraphics[width=1\linewidth]{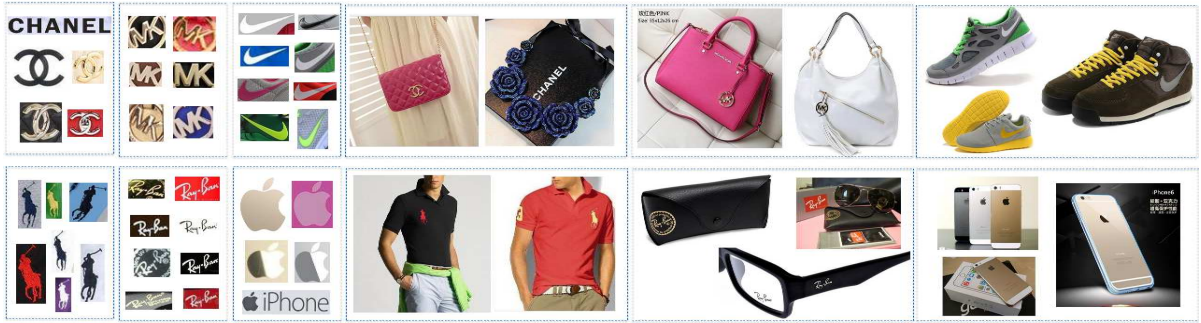}\vspace{-0.5in}
\end{center}%
\caption{Examples of Brands and logo images from real-world product images}
\vspace{-0.2in}
\end{figure*}%

The rest of this paper is organized as follows. Section \ref{section:problem formulation} presents the problem formulation of logo detection and brand recognition tasks from real-world product images. Section \ref{section:database} introduces the proposed ``LOGO-Net" database as well as presents the deep logo detection framework using the emerging deep region-based convolutional networks techniques. Section \ref{section:experiment} presents our empirical studies. Section \ref{section:conclusion} concludes this work.








\section{Problem Formulation}\label{section:problem formulation}

\subsection{Logo Detection}\label{section:Logo Detection}
In general, logo detection can be viewed as a special case of generic object detection in computer vision \cite{papageorgiou1998general,viola2001robust,russakovsky2014imagenet}, which is often more challenging than generic object recognition/classificatin tasks. Logo detection aims to detect logo instances of some pre-defined logo classes in digital images or videos. Similar to a generic object detection task, given an input image, a logo detection method not only needs to indicate if a logo is found in the image, but also needs to report the locations of the detected logo object instances/regionds found in the image.

Despite being a special case of generic object detection, logo detection from real-world product images (e.g., online shopping portals like Taobao.com or Aliexpress.com) is very challenging for several reasons. First of all, a logo instance occurring in a real product image can be extremely small. Second, a ``rigid" logo instance (in its original form) can become non-rigid when it is embedded in a natural product image, e.g., a logo occurring in a clothing image. Last but not least, logo instances of the same logo class occurring in real-world product images may differ very much due to a variety of changes, such as sizes, rotations, transformations, lighting, coloring, and occlusion, etc.


\subsection{Brand Recognition}\label{section:Brand Recognition}

Brand recognition aims to recognize the brand names of products in a real-world product image. From a machine learning and pattern recognition perspective, brand recognition is essentially a multi-class image classification task, where an input product image is classified into one of multiple pre-defined brand categories. The techniques for brand recognition from product images have many important applications, such as intellectual property protection in e-commerce, tracking brand-specific products for business intelligence, and visual online advertising, etc.

Although it is viewed as a multi-class image classification task, brand recognition cannot be solved directly by applying traditional image recognition techniques that simply do classification based on the visual contents of the whole product image. This is because the same brand can have multiple kinds of products (e.g., bags, shoes, or shirts, etc), and thus visual contents of product images from the same brand could be completely different.

To tackle the above challenge, we propose to explore logo detection techniques for brand recognition tasks. By detecting the appearance of logo objects related to a certain brand in a product image, one can solve the brand recognition task in an effective way. Therefore, the challenge of brand recognition can be reduced into solving a logo detection task from real product images. Finally, we note that a single brand can consist of multiple logo classes.

\begin{figure*}[htb]
\vspace{-0.3in}
\begin{center}
\hspace{-0.3in}\includegraphics[width=1.05\linewidth]{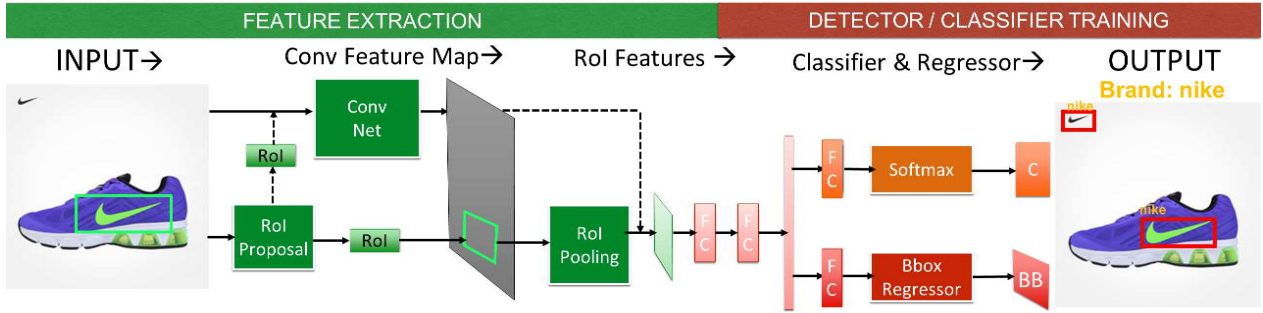}\vspace{-0.5in}
\end{center}%
\caption{DeepLogo-DRCN: {\small The proposed architecture of Deep Logo Detection using Deep Region-based Convolutional Networks (DRCN) techniques. Given an input image, the region proposal generation step yields a set of region of interests (RoIs) using an efficient implementation of Selective Search (SS)~\cite{uijlings2013selective}. The RoIs are then input into a fully convolutional neural network, in which each RoI is pooled into a fixed-size feature map and then mapped to a feature vector by fully connected layers (FCs), which is followed by training the final object classifiers and bounding box regressors. For each RoI, the network yields two output vectors: softmax probabilities and per-class bounding-box regression offsets. The overall architecture typically can be trained in an end-to-end approach, where the DRCN technique can be any existing R-CNN variants for generic object detection. For example, the solid-line process includes several emerging fast variants of R-CNN, such as fast R-CNN (FRCN)~\cite{girshick15fastrcnn} and SPPnet~\cite{he2014spatial}, while the dashline process represents the classical R-CNN~\cite{girshick2014rich}.}}\label{fig:framwork}
\vspace{-0.1in}
\end{figure*}%

\subsection{Deep Learning Framework}\label{section:framework}

Deep learning has achieved promising results in varied object detection tasks recently \cite{sermanet2013overfeat,girshick2014rich,erhan2014scalable}. One of the most successful deep learning paradigms for object detection is the series of region-based convolutional neural networks (R-CNN) \cite{girshick2014rich,he2014spatial,girshick15fastrcnn,DBLP:conf/cvpr/OuyangWZQLTLYWL15,ren2015faster}, which had obtained state-of-the-art results in many object detection benchmarks \cite{russakovsky2014imagenet}. Motivated by the successes of R-CNN related techniques for generic object detection, we propose a deep logo detection framework for brand recognition from real-world product images by exploring a family of state-of-the-art deep region-based convolutional networks (DRCN) techniques.

Figure \ref{fig:framwork} shows the proposed DeepLogo-DRCN framework for logo detection and brand recognition from product images. Specifically, given an input image, the region proposal step yields a set of region of interests (RoIs) using an efficient implementation of Selective Search (SS)~\cite{uijlings2013selective}. The RoIs are then input into a fully convolutional neural network (CNN), in which each RoI is pooled into a fixed-size feature map and then mapped to a feature vector by fully connected layers (FCs), followed by training the object classifiers and bounding box regressors. For each RoI, the network yields two kinds of outputs: softmax probabilities and per-class bounding-box regression offsets. The overall architecture can be trained end-to-end. The DeepLogo-DRCN framework can take any recent DRCN algorithms for generic object detection, e.g., traditional R-CNN \cite{girshick2014rich} (as shown by dashline), fast R-CNN (FRCN)~\cite{girshick15fastrcnn} and SPPnet~\cite{he2014spatial}, etc.

\section{LOGO-Net: large-scale logo image database}\label{section:database}
In this paper, we present ``LOGO-Net" --- a large-scale logo image database to facilitate the research of logo detection and brand recognition tasks with emerging deep learning techniques. Constructing such a large-scale real-world logo image database is very challenging, time-consuming, and costly. In the following, we discuss some major tasks and efforts in constructing our logo image database, especially data collection and data annotation tasks. We will then present two versions of datasets in our current LOGO-Net database, including a medium-scale dataset to ease the evaluations of varied settings, and a large-scale dataset. Finally, we compare our database against some of existing publicly available logo datasets.

\subsection{Product Image Collection}\label{section:collection}


The first task of LOGO-Net is to construct a list of brands and their associated popular logos according to the application needs. After building the list, for each brand and logo, we crawled their related product images from two online retail marketplaces: \url{www.taobao.com} --- the world's largest marketplace for online shopping targetted at Chinese consumers, and \url{www.aliexpress.com} --- a global retail marketplace targeted at consumers worldwide. All the product images crawled in our database were publicly available in the online marketplaces. In our database, each brand may have different number of images and each logo class may have different number of logo object instances. Our principle in the data collection is to ensure that each logo class at least has a minimal number of logo object instances for deep-learning research purposes.

\subsection{Logo Object Annotation}\label{section:annotation}

\begin{figure}[htb]
\vspace{-0.2in}
\begin{center}
\includegraphics[width=0.8\linewidth]{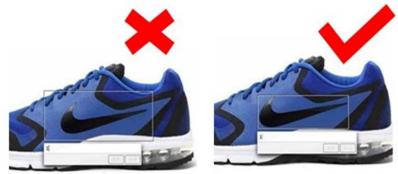}\vspace{-0.4in}
\end{center}
\caption{Instruction example of logo object annotation. The left-hand side is rejected due to too loose bounding box.}\label{fig:data-annotation-example}
\vspace{-0.2in}
\end{figure}

One of the most time-consuming and costly processes in constructing the LOGO-Net database is to annotate logo objects from the collected product images. For each product image, a human annotator needs to identify the logo objects, annotate the bounding box of each logo object, and then tag it with the corresponding logo class id. Figure \ref{fig:data-annotation-example} shows examples of logo object annotation on product images.

\begin{table}[H]
\vspace{-0.05in}
\begin{center}
\begin{tabular}{|l|c|c|}
\hline
Statistics & Logos-18&Logos-160\\
\hline
\# brands classes   &10     &100\\
\# logo classes     &18     &160\\
\# images           &8460   &73414\\
\# logo objects     &16043  &130608\\
mean image width    & 564 pixels   &687 pixels\\
mean image height   & 498 pixels   &707 pixels\\
\hline
\end{tabular}
\end{center}\vspace{-0.2in}
\caption{Dataset summary of Logos-18 and Logos-160}\label{table:18 and 160}
\vspace{-0.1in}
\end{table}

\begin{figure*}[htpb]
\vspace{-0.1in}
\begin{center}
 \includegraphics[width=1\linewidth]{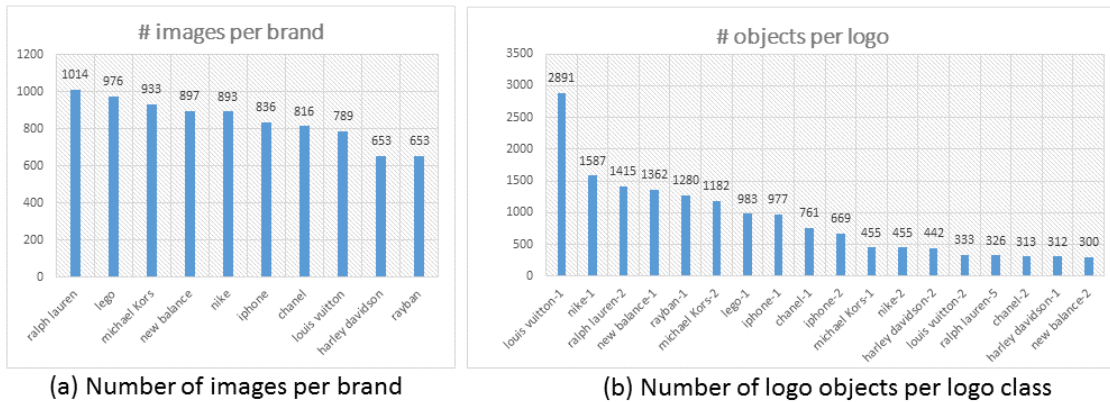}\vspace{-0.5in}
\end{center}
\caption{The statistics of numbers of images per brand and logo objects per logo class in the ``logos-18" dataset}\label{fig:logo18num}
\vspace{-0.1in}
\end{figure*}

\begin{figure*}[htb]
\vspace{-0.1in}
\begin{center}
\includegraphics[width=1\linewidth]{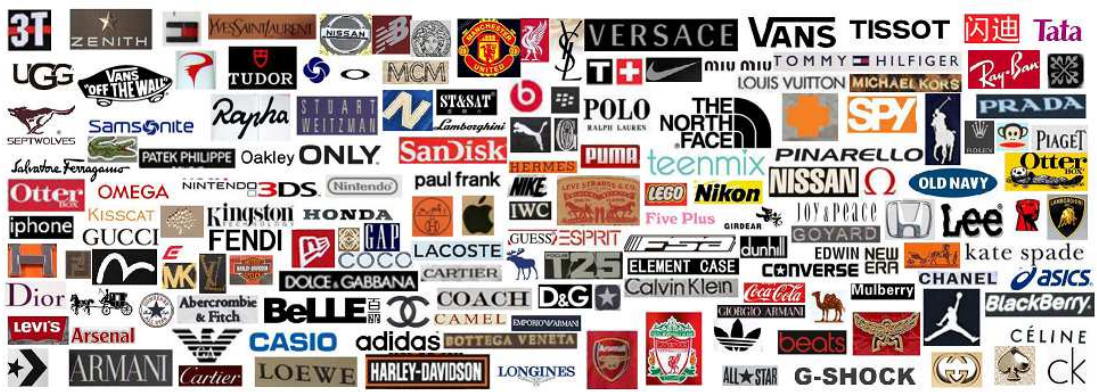}\vspace{-0.6in}
\end{center}%
\caption{Examples of logo images from the LOGO-Net database}\label{fig:example}
\vspace{-0.1in}
\end{figure*}%

\subsection{LOGO-Net Datasets: Logos-18 vs. Logos-160}\label{section:18 and 160}

\begin{figure*}[htb]
\vspace{-1in}
\hspace{-0.5in}\scalebox{1.0}{
\includegraphics[width=1.2\linewidth]{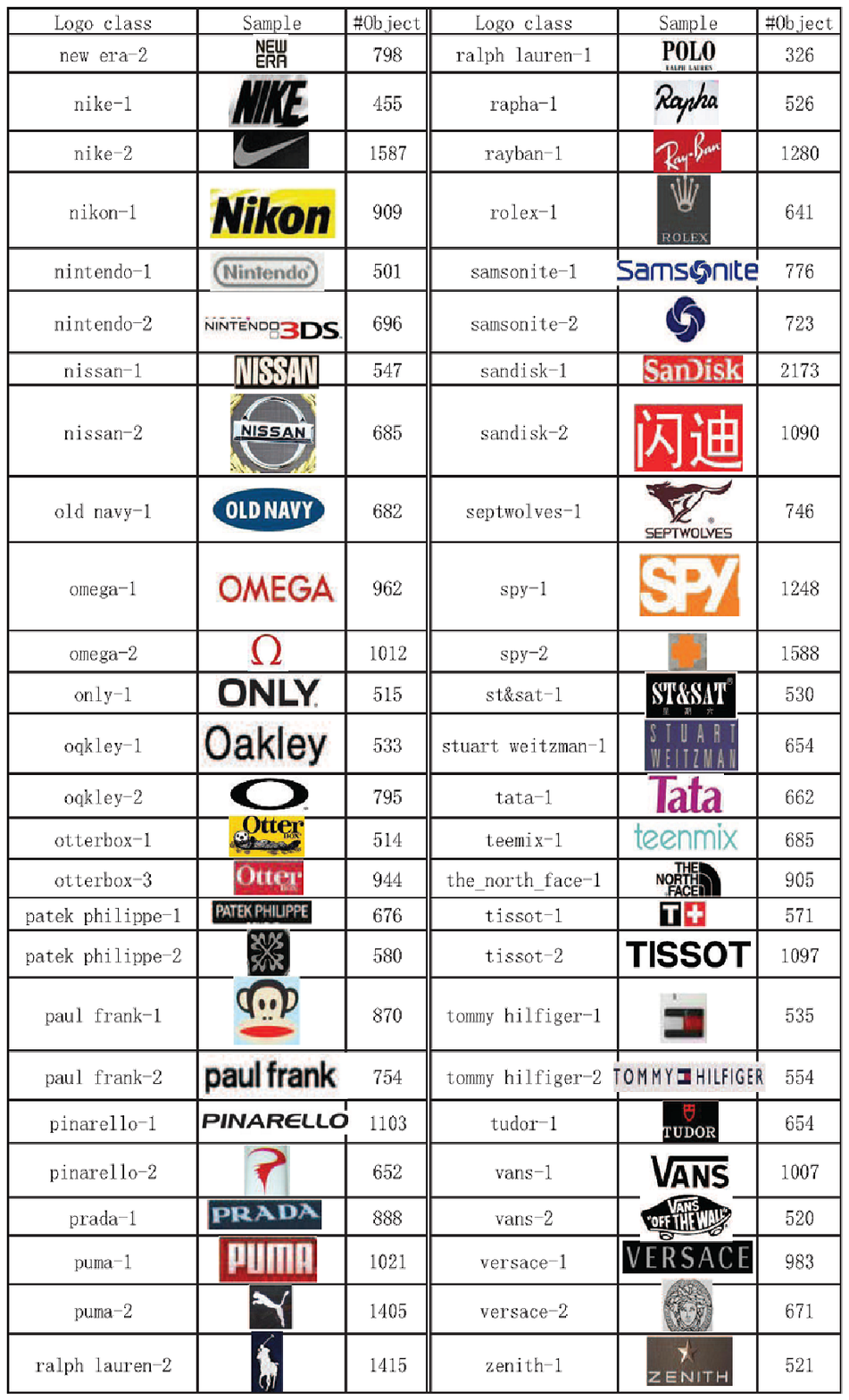}
}
\vspace{-1in}
\caption{List of 160 logo classes and their logo object statistics in our Logos-160 dataset}\label{figure:dataset-160-logo}
\vspace{-0.1in}
\end{figure*}

Object detection is computationally very intensive for both training and test. To facilitate research, we design two datasets of different scales: Logos-18 versus Logos-160.
Table \ref{table:18 and 160} shows the dataset summary of LOGO-Net. More details are given in the appendix (Table \ref{table:dataset-160-brand} and Figure \ref{figure:dataset-160-logo}).

\subsection{Comparison to Other Logo Data Sets}\label{section:compare}

We compare our Logos-18 and Logos-160 datasets with some other publicly available logo datasets. Table \ref{table:compare} gives the summary of the dataset comparisons.

\begin{table}[htb]
\begin{small}
\begin{center}
\begin{tabular}{|l|c|c|c|c|}
\hline
Dataset	&\#Image	&\#Logo&	\#Brand &\#Logo Object\\
\hline
Logos-160&73414&160&100&130608\\
\hline
Logos-18&	8460	&18&	10&	16043\\
\hline
BelgaLogos&	10000&	37&	37&	2695\\
\hline
FlickrLogos-27&	1080	&27&	27&	4671\\
\hline
FlickrLogos-32&	8240	&32&	32&	5644\\
\hline
\end{tabular}
\end{center}
\end{small}\vspace{-0.1in}
\caption{Comparisons of existing logo datasets with Logos-18 and Logos-160. Note that BelgaLogos, FlickrLogos-27 and FlickrLogos-32 contain many non-logo images.}\label{table:compare}
\vspace{-0.1in}
\end{table}

\textbf{BelgaLogos Dataset}\cite{belgalogos2009} It contains 37 logo categories and a total of 10,000 images, in which the maximum height or width of each image has been re-sized to 800 pixels. However, the total of logo object instances is less than 3000.

\textbf{FlickrLogos-27 Dataset}\cite{logo-icmr2011} It has only 27 logo categories and a total of 1080 images, and each logo class has less than 50 images on average.

\textbf{FlickrLogos-32 Dataset} \cite{RombergICMR2011} It has 32 logo categories and a total of 8240 images. Similar to the FlickrLogos-27 Dataset, the number of images for each logo class is small, which is less than 70 images per logo class.

Unlike the existing datasets, the LOGO-Net database has a much larger scale in terms of both total number of logo objects and average number of logo objects per class, which is important and critical to explore any data-driven machine learning techniques for logo detection and recognition. Figure \ref{fig:logo18num} shows some detailed statistics of numbers of images per brand and logo objects per logo class in the logos-18 dataset, and Figure \ref{fig:example} shows some examples of logos in our LOGO-Net database. More details about the database can be found in the appendix section.


\begin{table*}[htb]
\vspace{-0.1in}
\centering
\begin{tabular}{l|c|c|c|r|r|r}
\hline
Algorithm(model) & mAP(\%)   & Accuracy (\%)   & AUC (\%) & total train time & test time / image & GPU memory\\\hline
\hline
RCNN(CaffeNet)      &   69.1       &    95.2       &      95.3       &       2444 (min)      &       20886(ms)  &2.39 (GB)\\
RCNN(CaffeNet-w/o-ft)&  55.1         &     86.5     &      86.4       &      1549 (min)       &       20881(ms)       &2.39 (GB)\\\hline
FRCN(CaffeNet)      &   58.8    &   93.2    & 	  92.0    &    147 (min) & 448 (ms)   &1.67 (GB)\\
FRCN(VGG1024)       &   59.8    &   94.8    &     93.6    &    253 (min) & 540 (ms)   &3.04 (GB)\\
FRCN(VGG16)         &   61.4    &   94.7    &     93.2    &   1312 (min) & 859 (ms)  &10.86 (GB)\\\hline
SPPnet(ZF-w/o-bb)  &   54.5    &   92.5     &   92.3          &    707 (min) & 968 (ms)  &2.21 (GB)\\
SPPnet(ZF)         &   59.1    &    92.5       &      92.3       &    749 (min) & 1199 (ms)  &2.21 (GB)\\
\hline
\end{tabular}\vspace{-0.1in}
\caption{{\bf Logos-18 test set} Logo Detection and Brand Recognition Results by DeepLogo-DRCN with different algorithms. Note that the test time cost includes the average region proposal time (310ms) by SS which is the same for all the algorithms.Here RCNN(CaffeNet-w/o/-ft) means RCNN using ImageNet-pretrained CaffeNet for feature extraction without fine-tuning and SPPnet(ZF-w/o-bb) means SPPnet using ImageNet-pretrained ZF Net without bounding box regression.}
\label{table:Logo18 test}
\end{table*}%

\begin{table*}[htb]
\vspace{-0.1in}
\centering
\begin{tabular}{l|c|c|c|r|r|r}
\hline
Alg(model) & mAP(\%)   & Accuracy (\%)   & AUC(\%) & total train time & test time / image & GPU memory\\\hline
\hline
RCNN(Caffenet) &      69.9     &   90.0        &      89.6       &    8783 (min)  &  27273 (ms)      & 3.74(GB)\\\hline
FRCN(CaffeNet)      &    61.0       &    81.6       & 	  82.6        &    169 (min)    & 685 (ms)   &1.71 (GB)\\
FRCN(VGG1024)       &  60.3         &  81.3         &     82.3        &    283 (min)    & 752 (ms)   &3.05 (GB)\\
FRCN(VGG16)         &    65.8       &    85.8       &       86.8      &    1362 (min)   & 1044 (ms)   &10.89 (GB)\\\hline
SPPnet(ZF-w/o-bb)\quad\quad  &     53.6      &    81.8       &     82.0        &  1360  (min) & 1218 (ms)    &3.63 (GB)\\
SPPnet(ZF)         &    58.1      &     81.8      &    82.0        &   1494 (min) & 1639 (ms)    &3.63 (GB)\\
\hline
\end{tabular}\vspace{-0.1in}
\caption{{\bf Logos-160 test set} Logo Detection and Brand Recognition Results by DeepLogo-DRCN with different algorithms.
Note that the test time cost includes the region proposal time (467ms) by SS which is the same for all the algorithms.}
\label{table:Logo160 test}
\vspace{-0.2in}
\end{table*}%

\section{Experiments}\label{section:experiment}
\subsection{Experimental Setup}\label{section:setup}

To enable the benchmark research, we follow standard competition setups for data partitions in our experiments. Specifically, for Logos-18, we randomly divide the dataset into three parts: 50\% for training, 20\% for validation, and 30\% for test. Similarly, we also divide the Logos-160 dataset into three parts: 20\% for training, 20\% for validation, and 60\% for test. We purposely set less training data for the Logos-160 task to make it more challenging and realistic in real-world settings as requiring too much training data is not so impractical to scale.

We develop the proposed DeepLogo-DRCN scheme for logo detection and brand recognition by exploring several state-of-the-art Deep Region-based Convolutional Networks (DRCN) techniques for object detection, including
\begin{itemize}\vspace{-0.1in}
\item{RCNN \cite{girshick2014rich}}\footnote{\url{https://github.com/rbgirshick/rcnn}} is the most popular and widely used deep learning framework for object detection by combining region proposal (e.g., selective search~\cite{uijlings2013selective}) with convolutional neural networks (CNNs);\vspace{-0.1in}
\item{FRCN \cite{girshick15fastrcnn}}\footnote{\url{https://github.com/rbgirshick/fast-rcnn}} is a Fast R-CNN framework for object detection with deep region-based convolutional neural networks, which significantly improves computational efficiency of traditional R-CNN methods. \vspace{-0.1in}
\item{SPPnet \cite{he2014spatial}}\footnote{\url{https://github.com/ShaoqingRen/SPP_net}} is another variant of fast R-CNN by improving computational efficiency and exploring spatial pyramid pooling strategies.
\end{itemize}

Another important step in the DeepLogo-DRCN framework is the region proposal solution, which affects both logo detection quality and computational efficiency. In our approach, we employs the Selective Search (SS) method~\cite{uijlings2013selective} which has been shown as the state-of-the-art region proposal that often achieves the best quality. However, the original SS implementation is notably slow particularly when dealing with a large image. Instead of exploring other fast region proposal techniques~\cite{zitnick2014edge}
which often sacrifice quality, we have done a fast implementation of the SS method, which yields almost the same quality as the original SS implementation but is much faster in an order of magnitude. In our experiments,
we choose the number of RoIs to 2000 for all schemes according to the validation set in order to balance the trade-off between quality and efficiency.

For parameter settings, we choose the parameters of different algorithms using the same validation set. We set the number of fine-tuning iterations to 50,000 for all schemes whenever applicable, and the default threshold of Intersection over Union (IoU) to 0.5 for validating object bounding boxes. We will evaluate how different settings (e.g., the number of RoIs, the number of fine-tuning iterations, IoU, etc) affect the performance in parameter sensitivity section.

For performance evaluation metrics, we adopt the widely used mean Average Precision (mAP) for evaluating logo object detection tasks. For brand recognition tasks, we adopt the standard metrics for object recognition, including classification accuracy and Area under the ROC curve (AUC). The experiments were conducted in a GPU cluster with the NVIDIA high-end Tesla K80 GPU (2x Kepler GK210, 2496 cores per GPU, 12GB memory per GPU).




\begin{table*}[htb]
\scriptsize
  \centering
\renewcommand{\arraystretch}{1.3}
    \scalebox{0.9}{
    \begin{tabular}{l|c|*{18}{c}}
    \hline
    {\scriptsize Algorithm(model)}& {\scriptsize mAP} & cls1 & cls2 & cls3 & cls4 & cls5 & cls6 & cls7 & cls8 & cls9 & cls10 & cls11 & cls12 & cls13 & cls14 & cls15 & cls16 & cls17 & cls18 \\\hline
    \hline
    { RCNN(CaffeNet)} & {\footnotesize 69.1}  & 78.9 & 57.2    & 58.3  & 56.7  & 79.9  & 68.6  & 99.6    & 50.8  & 60.8  & 62.8  & 54.0  &89.2  & 52.7  & 68.1  & 79.5  &90.0  & 67.3  & 69.5  \\

    { FRCN(VGG16)}  & {\footnotesize 61.4}   & 75.2  & 50.8  & 57.0    & 56.2  & 69.4  & 67.2  & 99.5  & 42.5  & 47.0    & 46.1  & 28.2  & 89.1  & 44.6  & 58.6  & 77.2  & 82.2  & 48.7  & 64.9 \\

    { FRCN(VGG1024)} & {\footnotesize 59.8} &74.6  & 46.1  & 58.1  & 53.8  & 74.4  & 73.2  & 99.6  & 43.1  & 39.2  & 47.9  & 20.4  & 88.8  & 37.0    & 55.4  & 74.6  & 82.0    & 46.1  & 62.1 \\

    { FRCN(CaffeNet)} & {\footnotesize 58.8} &69.0    & 43.2  & 55.7  & 48.6  & 69.4  & 71.5  & 99.7  & 46.7  & 41.5  & 47.4  & 27.2  & 88.9  & 39.4  & 48.6  & 71.5  & 78.7  & 46.8  & 63.5 \\

    { SPPNet(ZF Net)} & {\footnotesize 59.2} &69.6 & 49.3    & 54.4  & 50.9  & 70.2  & 73.8  & 90.9&  41.3  & 38.6  & 48.4  & 48.5  & 84.3  & 37.7  & 49.8    & 67.2  & 79.3    & 50.6  & 61.5   \\
    \hline
    \end{tabular}}
\renewcommand{\arraystretch}{1.0}
    \vspace{-0.1in}
    \caption{{\bf Logos-18 test set logo detection} results of average precision (\%). The notations from ``cls 1" to ``cls-18" denote chanel-1, chanel-2, harley davidson-1, harley davidson-2,	iphone-1, iphone-2, lego-1, louis vuitton-1, louis vuitton-2, michael Kors-1,	michael Kors-2,	new balance-1, new balance-2, nike-1, nike-2, ralph lauren-2, ralph lauren-1, rayban-1, respectively.
}
  \label{table:Detail mAP}
\vspace{-0.1in}
\end{table*}%

\begin{table*}[htb]
\footnotesize
  \centering
    \scalebox{0.95}{
    \begin{tabular}{l|c|*{10}{c}}
    \hline
    {\scriptsize Alg(model)}& {\scriptsize Acc} & bnd1 & bnd2 & bnd3 & bnd4 & bnd5 & bnd6 & bnd7 & bnd8 & bnd9 & bnd10  \\\hline
    \hline
    { RCNN(CaffeNet)}  & { 95.2}  & 91.1 & 88.1    & 96.8  & 100.0  & 93.7  & 91.0  & 99.3    & 98.2  & 97.7 &92.3   \\

    { FRCN(VGG16)}  & { 94.7}   & 91.9  & 88.1  & 95.3    & 97.6  & 93.3  & 89.6  & 99.3  & 97.8  & 97.0    & 94.9 \\

    { FRCN(VGG1024)} & { 94.8} &92.7  & 87.6  & 91.7  & 99.7  & 93.3  & 91.4  &98.9& 97.0  & 98.7  & 93.3   \\

    { FRCN(CaffeNet)} & { 93.2} &90.3    & 85.5  & 88.5  & 99.7  & 90.4  & 86.7  & 98.9  & 97.4  & 97.4  & 92.8   \\

    { SPPNet(ZF Net)}  & { 92.5} &81.9 &88.6    &91.3  & 99.7  & 93.7  & 84.2  & 97.8&  94.8  & 98.7  & 91.3   \\
    \hline
    \end{tabular}}
\vspace{-0.1in}
\caption{{\bf Logos-18 test set brand recognition accuracy} results(\%). ``bnd 1" to ``bnd-10" denote ``chanel", ``harley davidson",	``iphone", ``lego", ``louis vuitton", ``michael Kors", ``new balance", ``nike", ``ralph lauren", ``rayban", respectively.
}
  \label{table:Detail Accuracy}
\vspace{-0.2in}
\end{table*}%

\subsection{Main Results}\label{section:result}

Table \ref{table:Logo18 test} and Table \ref{table:Logo160 test} summarize the main results of logo object detection and brand recognition on the test sets by the proposed DeepLogo-DRCN with different algorithms and models. For each dataset (logos-18 or logos-160), all the models were trained on
the same training data set, and tested on the same test set. The validation set was only used for choosing key parameters of each scheme. Several observations can be drawn from the main experimental results.

First of all, among all the methods, RCNN(CaffeNet with fine-tuning) obtained the best logo detection and brand recognition results on the logos-18 dataset, while FRCN(VGG16) obtained the second best results on the logos-18 dataset. This seems a bit surprising as both FRCN and SPPnet have been reported with state-of-the-art results, if not better than, at least comparable to RCNN for generic object detection on PASCAL VOC object detection benchmarks. This is mainly because unlike PASCAL VOC datasets, our LOGO-Net database has many small logo objects for which FRCN and SPPnet might fail to detect if the convolutional feature map is not large enough. In contract, RCNN suffers less from this issue since RCNN first takes an RoI (a small region) and then resize it to a fixed size (essentially enlarged) before it is passed to the convolutional network.


Second, in terms of training time cost, we found that RCNN is the most computationally expensive among all the solutions. This is because RCNN has to repeatedly perform convolutional operations for each RoI with the original image, which can be somewhat redundant and thus very computationally expensive. By comparing FRCN and SPPnet, when using a simpler network (e.g., CaffeNet), FRCN can be trained several times faster than SPPnet (based on the Zeiler-Fergus's ZF-net) \cite{zeiler2014visualizing}. However, when using a very deep network (VGG16), FRCN is computationally more expensive than SPPnet with ZF-net. Moreover, by examining GPU memory cost during training, we found that FRCN consumes a large amount of GPU memory (e.g., more than 10GB) when training the very deep VGG16 network. This poses a very high requirement for the GPU hardware (e.g., requiring very high-end GPU, e.g., K80 in our cluster).

Moreover, by examining the test time cost which is critical when being deployed in real-world applications, we found that R-CNN takes more than 20 seconds for processing an image, which is an order of magnitude slower than the others. The poor prediction efficiency makes RCNN infeasible to be deployed in a real-world application which may need to deal with millions of product images daily.

Finally, Table \ref{table:Detail mAP} and Table \ref{table:Detail Accuracy} give the detailed results on logos-18 for specific logo detection and specific brand recognition accuracy, respectively. The appendix section also includes the detailed results on the Logos-160 dataset (Table \ref{table:Detail mAP-160} and Table \ref{table:Detail Accuracy-160}).




\vspace{-0.05in}
\subsection{Parameter Sensitivity}\label{section:parameter}

\vspace{-0.05in}
\subsubsection{Overview}\label{section:para overview}

For the proposed DeepLogo-DRCN, there are some key parameters that may significantly affect logo detection and brand recognition performance, including the number of regions of interests (RoIs),
the number of fine-tuning (ft) iterations, the amount of training data, the IoU setting, extra acceleration using SVD, etc. In this section, we evaluate how the performance is sensitive to each of these factors.

\vspace{-0.05in}
\subsubsection{Evaluation of Number of Regions of Interests}\label{section:RoI}

The number of RoIs (i.e., the number of bounding boxes) yielded by SS \cite{uijlings2013selective} plays a critical role in DeepLogo-DRCN, which affects both detection quality and recognition accuracy as well as computational efficiency. Computational cost is generally proportion to the number of RoIs. The higher the number of RoIs, the more computational costs for both training and test.
However, increasing the RoI size may not always guarantee a significant improvement of mAP.

\begin{figure*}[htb]
\vspace{-0.1in}
\hspace{-0.3in}
\begin{subfigure}{.35\textwidth}
  \centering
  \includegraphics[width=1.1\linewidth]{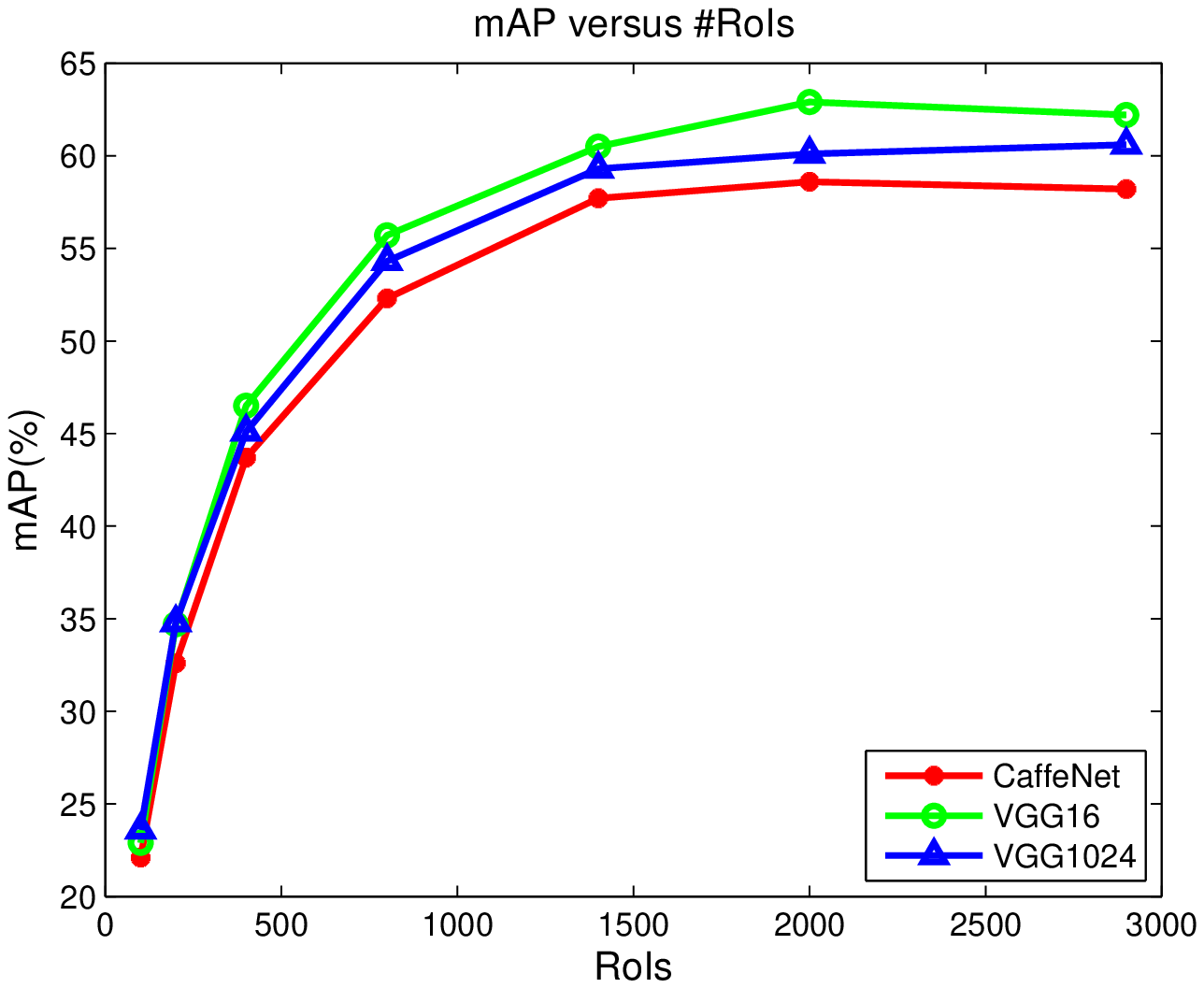}\vspace{-0.1in}
  \caption{mAP versus \#RoIs}
  \label{fig:nb-RoIs-a}
\end{subfigure}
\begin{subfigure}{.35\textwidth}
  \centering
  \includegraphics[width=1.1\linewidth]{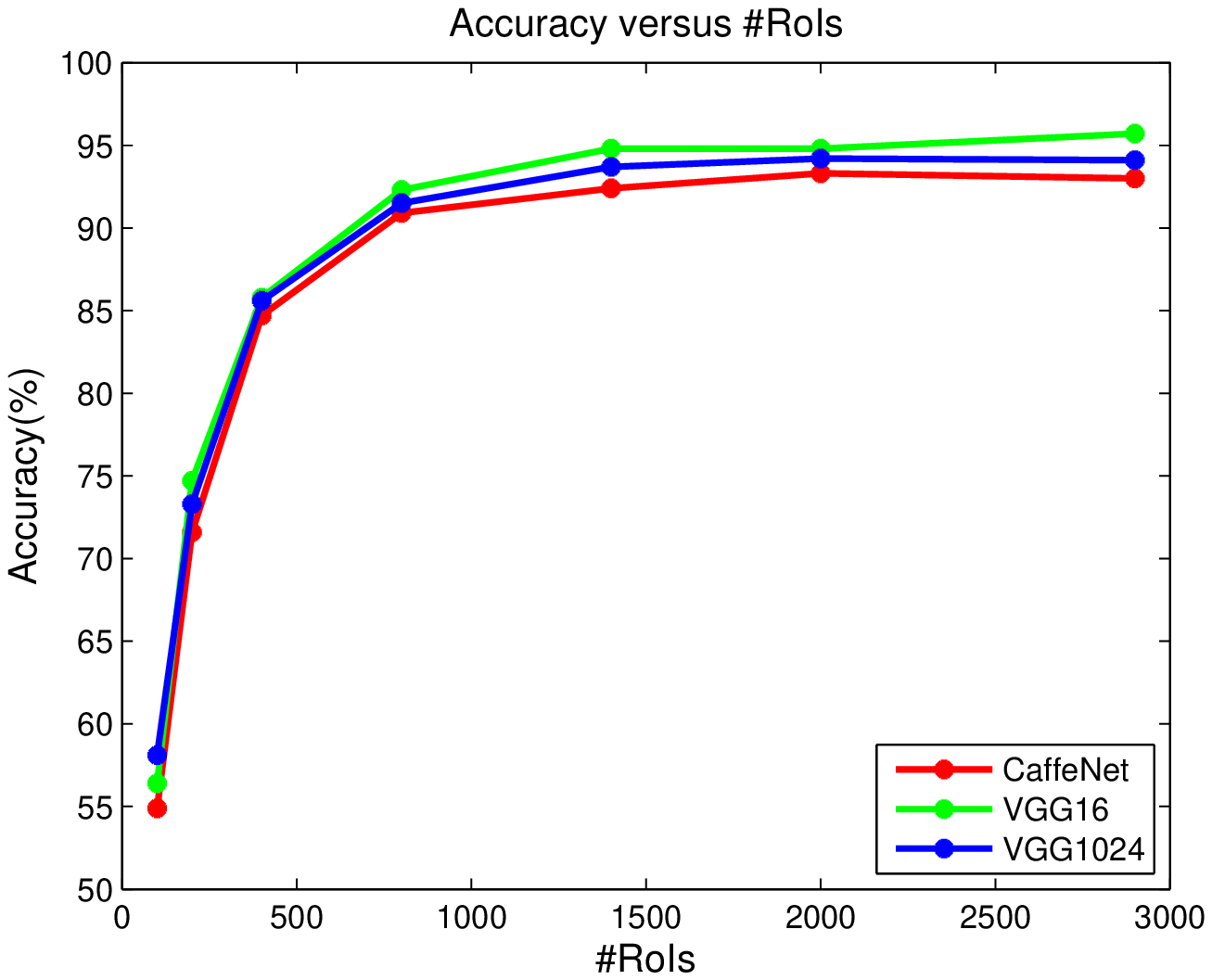}\vspace{-0.1in}
  \caption{Accuracy versus \#RoIs}
  \label{fig:nb-RoIs-b}
\end{subfigure}
\begin{subfigure}{.35\textwidth}
  \centering
  \includegraphics[width=1.1\linewidth]{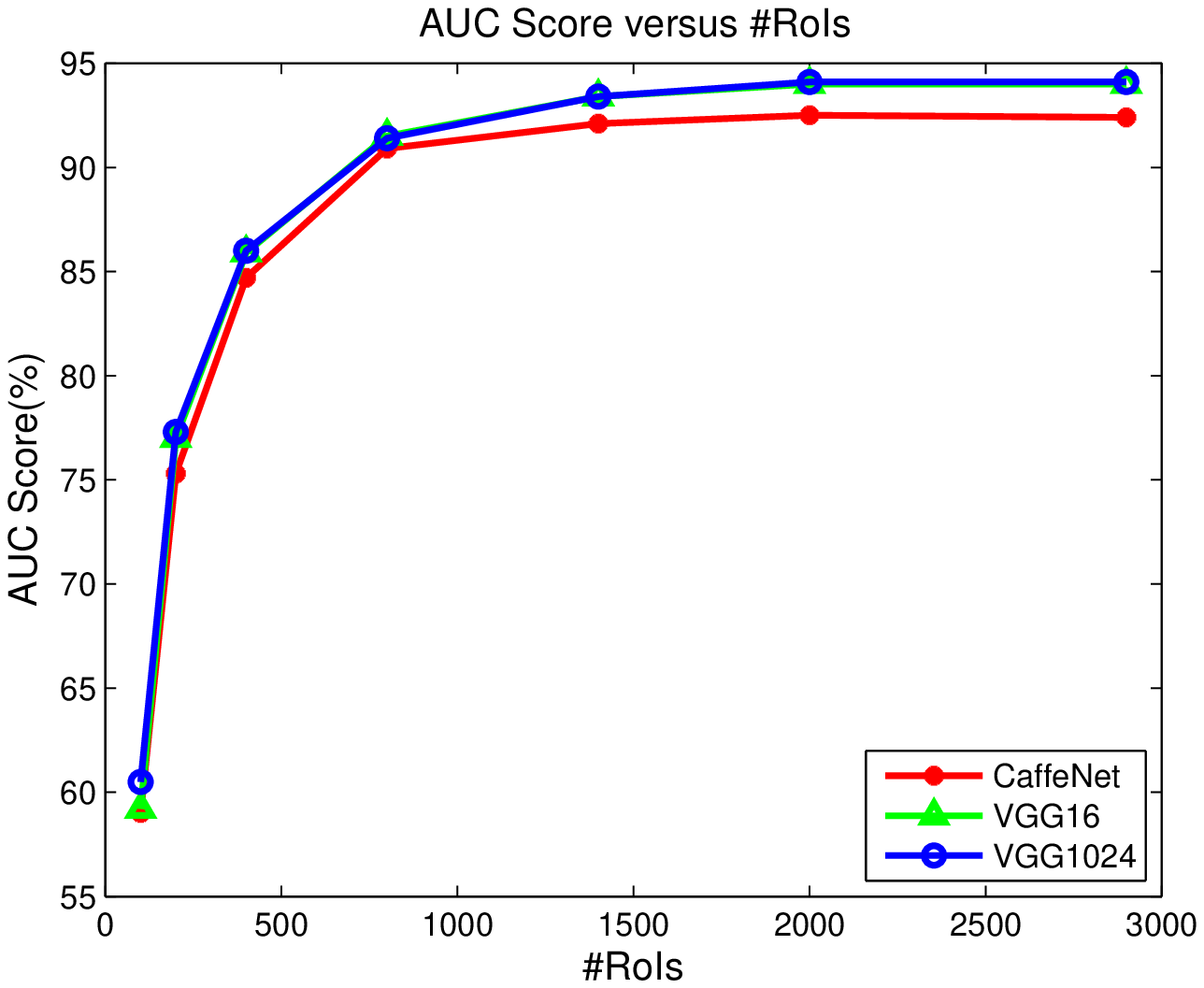}\vspace{-0.1in}
  \caption{AUC versus \#RoIs}
  \label{fig:nb-RoIs-c}
\end{subfigure}\vspace{-0.1in}
\caption{Evaluation of parameter sensitivity of \#RoIs (bounding boxes) for object detection (mAP) and brand recognition performance (accuracy and AUC). The logo object detection algorithm is based on DeepLogo-FRCN.}
\label{fig:nb-RoIs}
\end{figure*}

\begin{figure*}[htb]
\vspace{-0.1in}
\hspace{-0.3in}
\begin{subfigure}{.35\textwidth}
  \centering
  \includegraphics[width=1.1\linewidth]{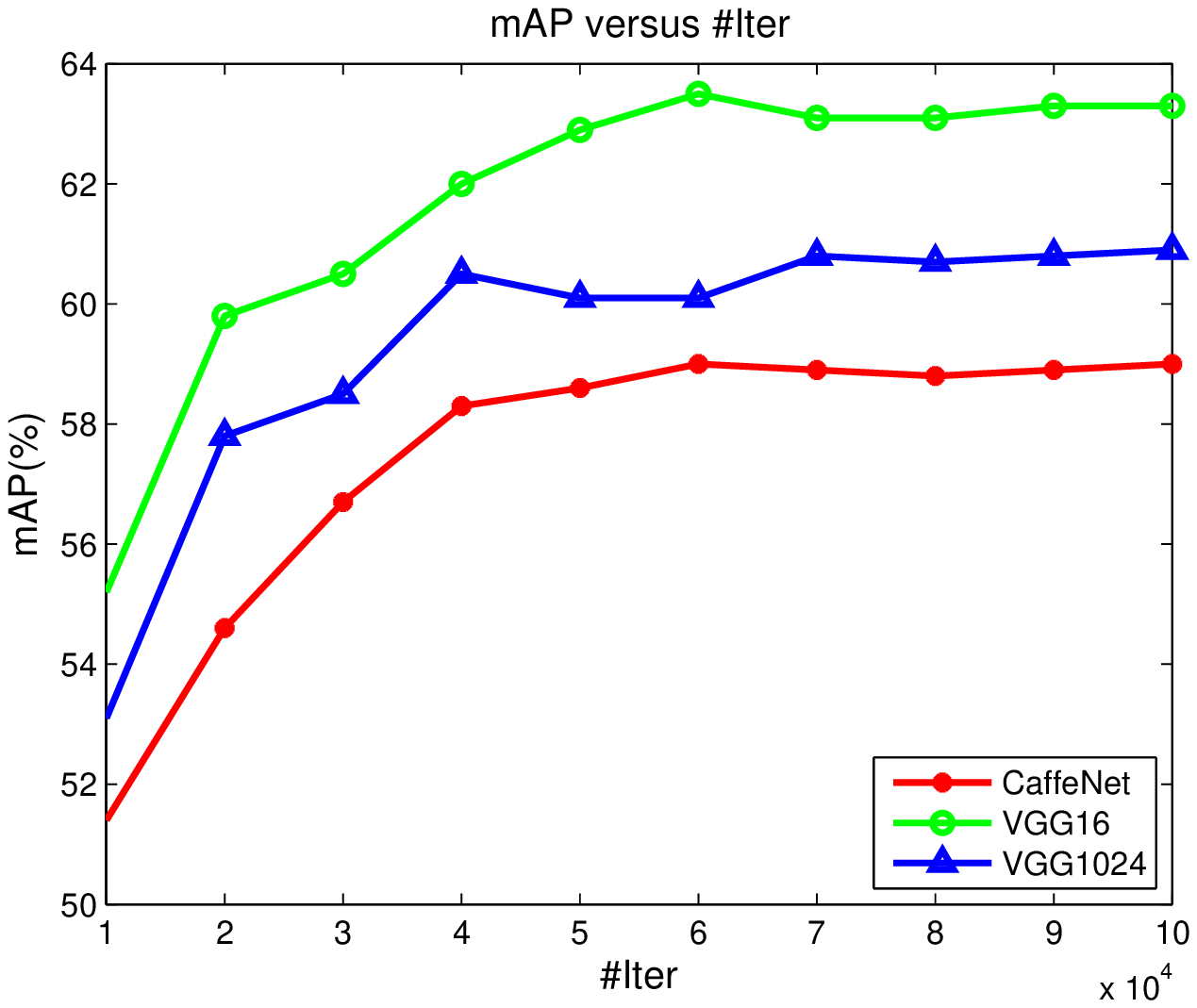}\vspace{-0.1in}
  \caption{mAP versus \#Iter}
  \label{fig:nb-Iter-a}
\end{subfigure}
\begin{subfigure}{.35\textwidth}
  \centering
  \includegraphics[width=1.1\linewidth]{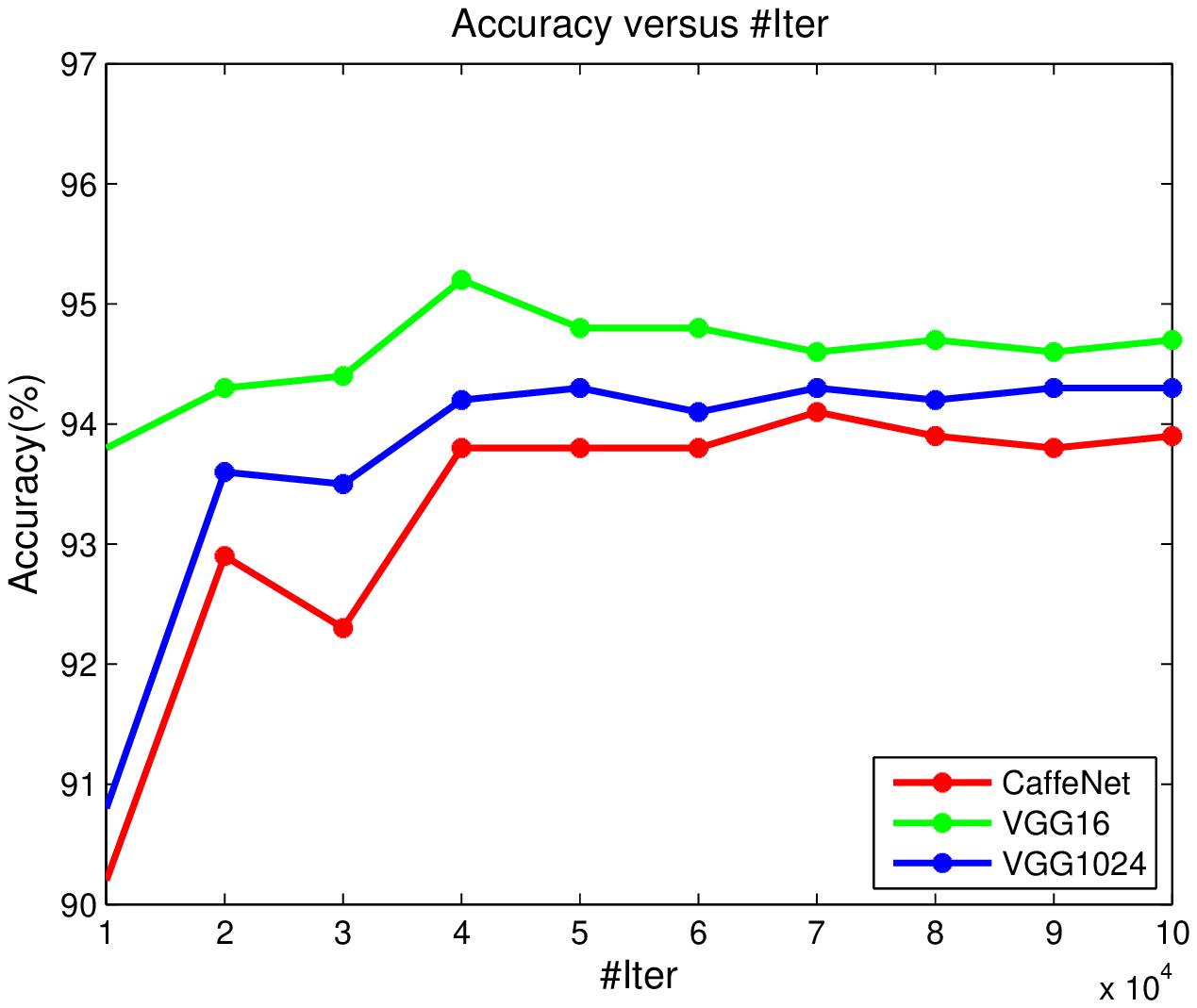}\vspace{-0.1in}
  \caption{Accuracy versus \#Iter}
  \label{fig:nb-Iter-b}
\end{subfigure}
\begin{subfigure}{.35\textwidth}
  \centering
  \includegraphics[width=1.1\linewidth]{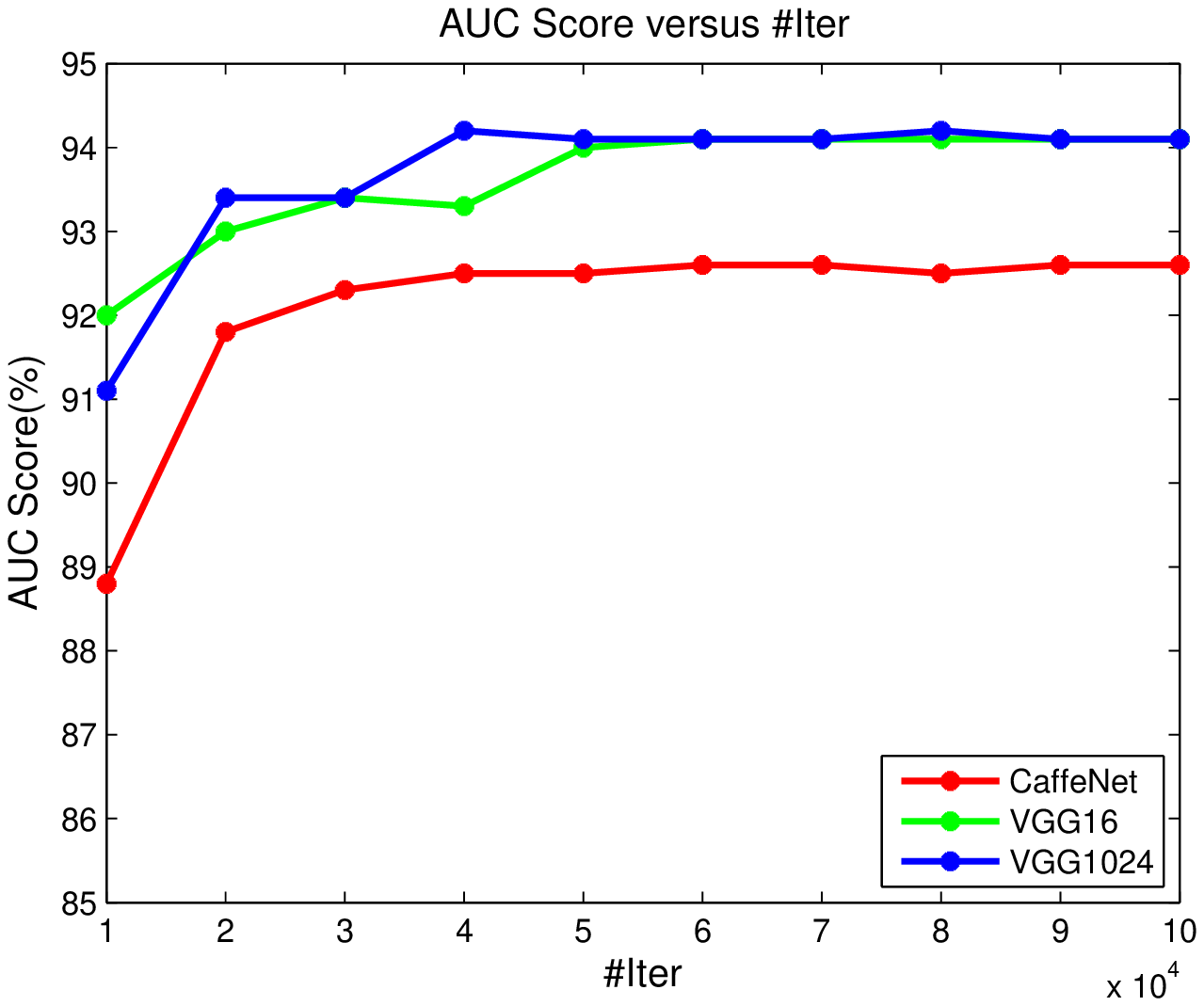}\vspace{-0.1in}
  \caption{AUC versus \#Iter}
  \label{fig:nb-Iter-c}
\end{subfigure}
\vspace{-0.1in}
\caption{Evaluation of parameter sensitivity of \#Iterations (fine-tuning) for object detection (mAP) and brand recognition performance (accuracy and AUC). The logo object detection algorithm is based on DeepLogo-FRCN.}
\label{fig:nb-Iter}
\vspace{-0.2in}
\end{figure*}

Figure \ref{fig:nb-RoIs} shows how the mAP performance on the validation set is changed with respect to different numbers of RoIs in the proposed DeepLogo-DRCN with FRCN using different CNN models.
From the results, we can see that when the number of RoIs is small (e.g., less than 1000), increasing the number of RoIs always leads to a considerable improvement of overall mAP. However, when it is large enough (e.g., 2000),
increasing it may only make a marginal improvement or even degrade the performance when it is too large (perhaps due to noise reasons). Thus, we found that setting the number of RoIs to 2000 is able to make a good tradeoff between logo detection efficacy and computational efficiency. Finally, by examining the brand recognition results, we found that as compared with mAP, both accuracy and AUC are less sensitive to the number of RoIs. When
the number of RoIs is large than 1500, both accuracy and AUC results are almost saturated.


\begin{figure*}[htb]
\hspace{-0.3in}
\begin{subfigure}{.35\textwidth}
  \centering
  \includegraphics[width=1.1\linewidth]{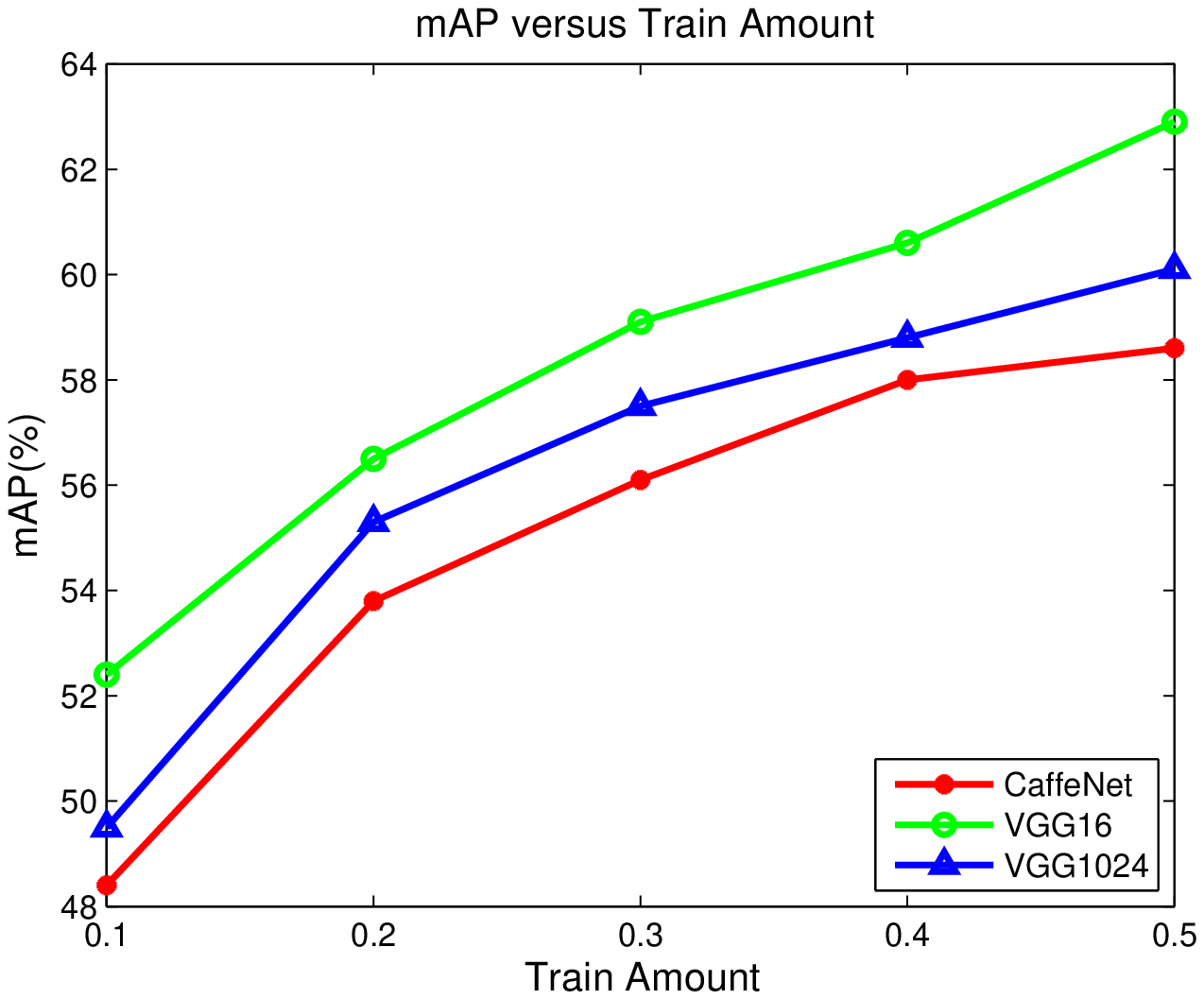}\vspace{-0.1in}
  \caption{mAP versus training data size}
  \label{fig:nb-Data-a}
\end{subfigure}
\begin{subfigure}{.35\textwidth}
  \centering
  \includegraphics[width=1.1\linewidth]{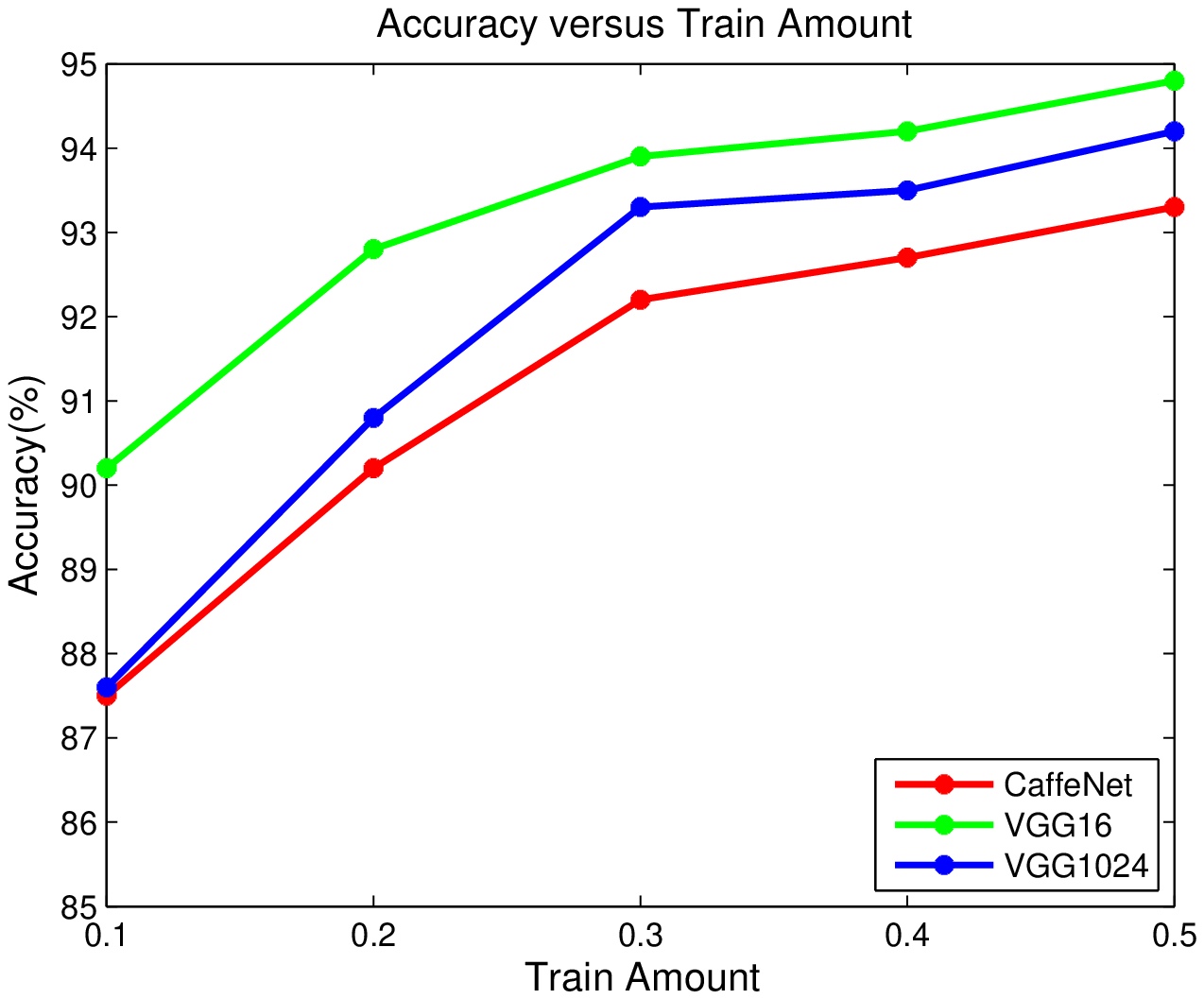}\vspace{-0.1in}
  \caption{Accuracy versus training data size}
  \label{fig:nb-Data-b}
\end{subfigure}
\begin{subfigure}{.35\textwidth}
  \centering
  \includegraphics[width=1.1\linewidth]{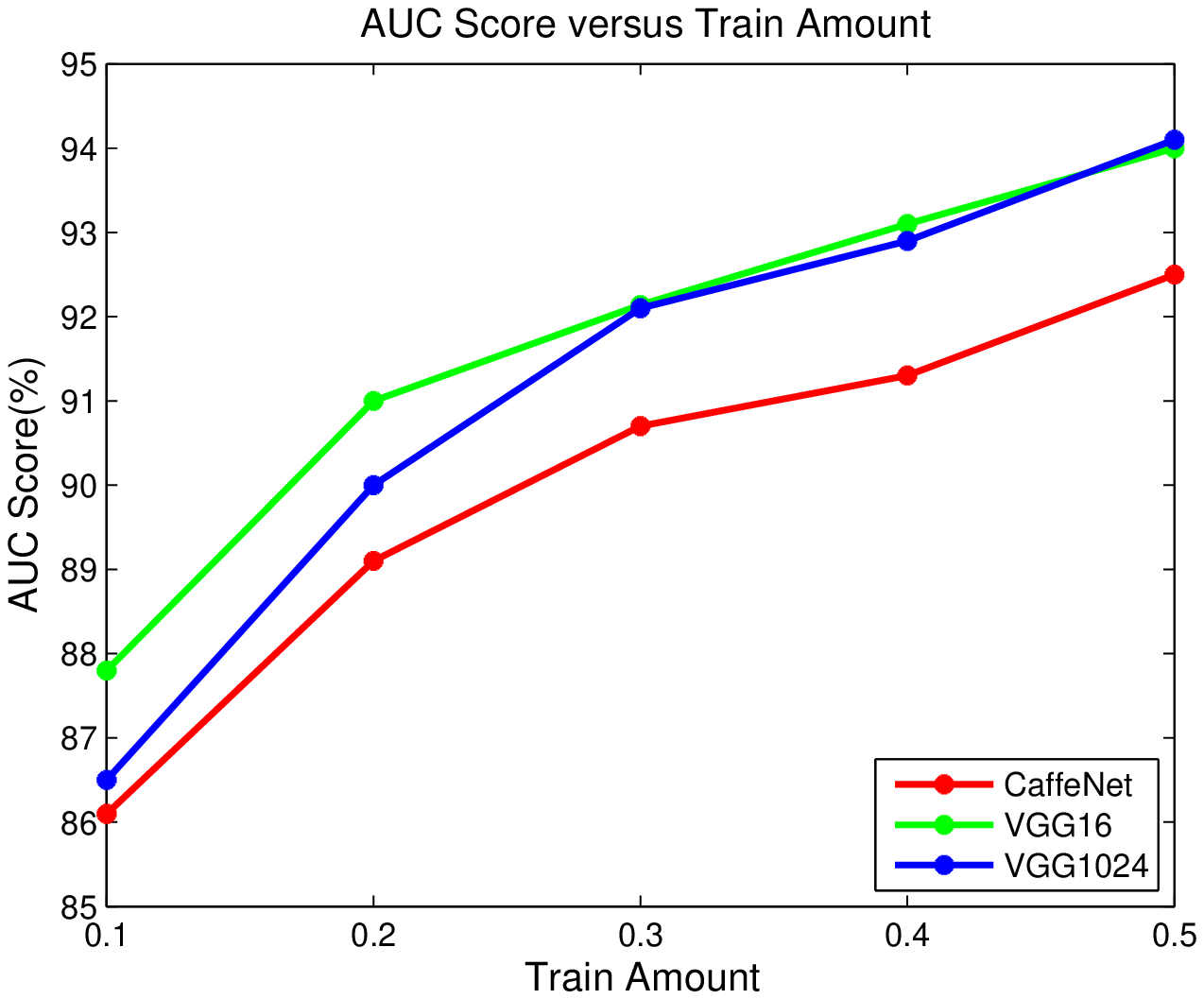}\vspace{-0.1in}
  \caption{AUC versus training data size}
  \label{fig:nb-Data-c}
\end{subfigure}
\vspace{-0.1in}
\caption{Evaluation of different amounts of training data for object detection (mAP) and brand recognition performance (accuracy and AUC). The logo object detection algorithm is based on DeepLogo-FRCN.}
\label{fig:nb-Data}
\vspace{-0.1in}
\end{figure*}

\vspace{-0.1in}
\subsubsection{Evaluation of Fine-Tuning (FT) Iterations}\label{section:FT}

The fine-tuning procedure is one of critical steps for ensuring the proposed DeepLogo-DRCN scheme can adapt the existing pre-trained CNN models on the logo image dataset domain. In general, we need a significantly large number
of fine-tuning iterations to ensure the proposed DeepLogo-DRCN is converged on the logo training data set. However, the training time cost grows linearly with the number of fine-tuning iterations. For a large-scale experiment, we need to set a proper number of fine-tuning iterations to balance the trade-off between efficacy and efficiency.

Figure \ref{fig:nb-Iter} shows how the overall logo detection (mAP) and brand recognition (accuracy) performances on the validation set are changed when increasing the fine-tuning iterations. We found that when setting it to about 50,000 iterations, the mAP performance will converge for most settings for FRCN. Finally, by examining the accuracy and AUC results of brand recognition, we found that the performances are almost saturated after 40,000 fine-tuning iterations.

\vspace{-0.1in}
\subsubsection{Evaluation of Training Data Sizes}\label{section:data}

For deep learning methods, training data amount affects considerably the resulting performance. In this experiment, we aim to examine
how the logo detection and brand recognition results were sensitive to different amounts of training data used for training the DeepLogo-DRCN scheme.

Figure \ref{fig:nb-Data} shows the evaluation results of object detection and brand recognition performances with respect to different amounts of training data. From the results, we can see that increasing the amount of training data generally gives a consistent improvement of both logo object detection and brand recognition performances. The improvement of mAP is particularly more significant while the improvement of brand recognition accuracy is relatively less obvious. This is primarily because classification accuracy is already very good and thus making a further improvement would be more difficult. This result also indicate classification accuracy may not be an ideal performance metric as compared with either mAP or AUC metrics.

\begin{figure}[htbp]
\vspace{-0.1in}
\begin{center}
   \includegraphics[width=0.8\linewidth]{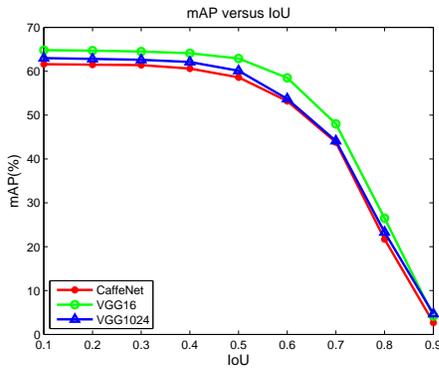}\vspace{-0.2in}
\caption{Evaluation of Intersection over Union (IoU) parameter settings for object detection performance (mAP).}\label{fig:IoU-1}
\end{center}
\vspace{-0.4in}
\end{figure}

\vspace{-0.1in}
\subsubsection{Evaluation of Intersection over Union (IoU)}\label{section:IoU}

The IoU threshold is a parameter to decide if a detected bounding box is overlapping enough with a target object. Setting a high IoU threshold requires a more precise localization of the detected object which is often done by the bounding box regression step. For brand recognition, it is less important for detecting precise bounding boxes of target logos. Figure \ref{fig:IoU-1} shows an evaluation of mAP with respect to different settings of IoU. We can see that when IoU is larger than 0.5, decreasing IoU only leads to a marginal improvement of mAP. However, when IoU is less than 0.5, increasing IoU can result in a significant drop of mAP.

\subsubsection{Evaluation of Acceleration by Truncated SVD}\label{section:SVD}

We realize the prediction time is crucial when applying DeepLogo-DRCN for real-world applications. To speed up the prediction of DeepLogo-DRCN,
we explore an SVD truncation based acceleration technique from \cite{girshick15fastrcnn}. The basic idea is to simplify the most intensive and repeatedly computation parts, i.e., the fully connected layers of DRCN,  using SVD-truncation based approximation. More details about SVD-approximation can be found in \cite{girshick15fastrcnn}.


%
\begin{table}[htb]

\begin{center}
\scalebox{1}{
\begin{tabular}{l|r|r|r|r}
\hline
{\small metric} & SVD(Y/N) &CaffeNet&VGG16& VGG1024\\
\hline
    & w/o SVD           & 58.8\% & 61.4\%&59.8\%\\
mAP & w/  SVD & 57.3\% & 60.8\%&59.6\%\\
    & drop in mAP      &  1.5\% & 0.6\% &0.2\%\\
\hline
    &w/o SVD    & 93.2\%& 94.7\%&94.8\%\\
Acc.        &w/  SVD & 92.4\% &94.6\%&94.5\%\\
& drop in acc.  &0.8\%&0.1\%&0.3\%\\
\hline
     &w/o SVD    & 92.0\%& 93.2\%&93.6\%\\
AUC        &w/  SVD & 91.5\%&93.0\%&93.5\%\\
& drop in AUC    &0.5\%&0.2\%&0.1\%\\
\hline
Test    &w/o SVD    & 0.137& 0.542&0.230\\
Time    &w/  SVD & 0.091 &0.391&0.149\\
& speedup      &33.6\%&27.9\%&35.2\%\\
\hline
\end{tabular}
}
\end{center}\vspace{-0.2in}
\caption{Evaluation of speedup gains obtained by SVD-based acceleration. Note that the above test time cost excludes the region proposal time cost by SS.}\label{table:SVD}
\end{table}

Table \ref{table:SVD} gives the speedup gains using the SVD-based acceleration. We can obtain about 30\% speedup while sacrificing only a minor drop in predictive performance. Specifically the gains obtained for deeper models are especially impressive. For example, for VGG16, we can obtain 28\% speedup in test time while only suffering no more than 0.2\% drop in both accuracy and AUC.



\section{Conclusions}\label{section:conclusion}

This paper presented ``LOGO-Net" --- a large-scale logo image database to facilitate large-scale deep logo detection and brand recognition from real-world product images. LOGO-Net consists of two datasets: (i)``logos-18": 18 logo classes, 10 brands, and 16,043 logo objects, and (ii) ``logos-160": 160 logo classes, 100 brands, and 130,608 logo objects. We discussed the challenges and solutions for constructing such a large-scale database, tackled the deep logo recognition and brand recognition tasks by exploring a family of emerging Deep Region-based Convolutional Networks (DRCN) techniques, and finally conducted an extensive set of benchmark evaluations.

\section*{Acknowledgements}

The authors would like to acknowledge the support from Dr Rong Jin at Alibaba Group to facilitate the research collaboration between Dr Hoi's team and Alibaba Group. The authors would like to acknowledge the fruitful discussions and help from researchers and engineers from Alibaba Group, especially the intern Mr Jiewei Luo.



\section*{Appendix: Details of Logo image database}

\begin{table*}[htbp]
  \footnotesize
  \centering
  \renewcommand{\arraystretch}{1.7}
   \scalebox{0.85}{
    \begin{tabular}{l|*{16}{c}}
    \hline
    Alg(Model) & cls1  & cls2  & cls3  & cls4  & cls5  & cls6  & cls7  & cls8  & cls9  & cls10 & cls11 & cls12 & cls13 & cls14 & cls15 & cls16 \\
    \hline
    \hline
    {RCNN(CaffeNet)} & 63.5  & 90.9  & 88.8  & 88.8  & 68.3  & 64.8  & 74.0  & 76.6  & 82.0  & 77.8  & 65.7  & 73.3  & 75.7  & 89.7  & 60.8  & 54.0  \\
    {FRCN(VGG16)} & 44.7  & 90.9  & 82.5  & 89.0  & 54.7  & 58.9  & 77.2  & 64.0  & 61.0  & 73.1  & 59.1  & 69.5  & 74.0  & 89.4  & 78.0  & 59.6  \\
    {SPPnet(ZF net)} & 43.8  & 90.3  & 67.0  & 87.3  & 30.4  & 49.9  & 57.6  & 73.9  & 61.6  & 61.1  & 36.5  & 62.6  & 59.2  & 85.7  & 64.5  & 36.7  \\
    \hline
    \hline
    Alg(Model) & cls17 & cls18 & cls19 & cls20 & cls21 & cls22 & cls23 & cls24 & cls25 & cls26 & cls27 & cls28 & cls29 & cls30 & cls31 & cls32 \\
    \hline
    \hline
    {RCNN(CaffeNet)} & 74.7  & 80.8  & 84.7  & 90.9  & 83.9  & 77.0  & 89.2  & 83.0  & 70.9  & 43.5  & 44.3  & 87.8  & 70.9  & 81.8  & 84.5  & 61.8  \\
    {FRCN(VGG16)} & 69.0  & 69.0  & 85.0  & 90.5  & 79.5  & 72.8  & 83.1  & 81.7  & 58.0  & 45.5  & 50.8  & 86.5  & 66.1  & 82.3  & 82.8  & 51.2  \\
    {SPPnet(ZF net)} & 46.1  & 59.5  & 82.2  & 83.7  & 70.8  & 79.5  & 80.1  & 53.8  & 43.3  & 31.4  & 43.6  & 79.4  & 61.1  & 80.9  & 73.9  & 26.8  \\
    \hline
    \hline
    Alg(Model) & cls33 & cls34 & cls35 & cls36 & cls37 & cls38 & cls39 & cls40 & cls41 & cls42 & cls43 & cls44 & cls45 & cls46 & cls47 & cls48 \\
    \hline
    \hline
    {RCNN(CaffeNet)} & 72.7  & 58.8  & 64.1  & 25.6  & 64.6  & 63.7  & 77.3  & 74.3  & 24.1  & 89.7  & 83.2  & 85.4  & 86.2  & 80.5  & 54.9  & 78.0  \\
    {FRCN(VGG16)} & 67.5  & 49.5  & 64.3  & 32.8  & 61.0  & 47.3  & 80.4  & 72.5  & 25.3  & 89.3  & 81.6  & 90.1  & 84.7  & 72.5  & 40.3  & 74.7  \\
    {SPPnet(ZF net)} & 56.3  & 35.4  & 59.2  & 25.8  & 45.6  & 34.6  & 75.7  & 56.8  & 25.4  & 88.8  & 74.0  & 83.4  & 78.8  & 62.0  & 29.0  & 65.3  \\
    \hline
    \hline
    Alg(Model) & cls49 & cls50 & cls51 & cls52 & cls53 & cls54 & cls55 & cls56 & cls57 & cls58 & cls59 & cls60 & cls61 & cls62 & cls63 & cls64 \\
    \hline
    \hline
    {RCNN(CaffeNet)} & 89.7  & 68.4  & 67.9  & 48.8  & 54.0  & 69.1  & 52.9  & 27.1  & 89.0  & 90.9  & 63.5  & 70.5  & 81.9  & 76.9  & 27.4  & 86.4  \\
    {FRCN(VGG16)} & 88.6  & 59.3  & 65.5  & 40.9  & 48.2  & 76.2  & 41.7  & 29.0  & 85.3  & 90.9  & 65.0  & 71.8  & 78.1  & 79.5  & 17.7  & 75.7  \\
    {SPPnet(ZF net)} & 78.9  & 46.6  & 57.0  & 36.1  & 43.3  & 66.9  & 31.5  & 24.4  & 76.1  & 90.9  & 57.9  & 62.6  & 71.4  & 70.9  & 22.5  & 78.5  \\
    \hline
    \hline
          & cls65 & cls66 & cls67 & cls68 & cls69 & cls70 & cls71 & cls72 & cls73 & cls74 & cls75 & cls76 & cls77 & cls78 & cls79 & cls80 \\
    \hline
    \hline
    {RCNN(CaffeNet)} & 51.5  & 45.1  & 63.6  & 43.7  & 41.2  & 41.0  & 64.1  & 55.5  & 83.2  & 53.6  & 63.7  & 69.3  & 45.5  & 70.7  & 61.4  & 32.5  \\
    {FRCN(VGG16)} & 31.6  & 50.6  & 58.6  & 43.0  & 42.3  & 35.0  & 55.7  & 34.9  & 84.9  & 59.0  & 62.8  & 63.3  & 47.3  & 71.4  & 53.6  & 26.0  \\
    {SPPnet(ZF net)} & 26.2  & 39.0  & 43.3  & 42.7  & 37.4  & 30.9  & 48.5  & 26.6  & 77.5  & 44.7  & 53.9  & 50.2  & 51.8  & 67.1  & 37.1  & 14.3  \\
    \hline
    \hline
          & cls81 & cls82 & cls83 & cls84 & cls85 & cls86 & cls87 & cls88 & cls89 & cls90 & cls91 & cls92 & cls93 & cls94 & cls95 & cls96 \\
    \hline
    \hline
    {RCNN(CaffeNet)} & 89.5  & 86.8  & 90.9  & 89.5  & 88.6  & 56.0  & 78.6  & 69.0  & 83.2  & 62.7  & 90.7  & 66.8  & 89.5  & 54.5  & 72.0  & 90.8  \\
    {FRCN(VGG16)} & 88.6  & 83.6  & 91.0  & 88.8  & 82.5  & 39.4  & 75.1  & 63.1  & 80.9  & 74.0  & 90.9  & 66.4  & 85.6  & 49.9  & 70.1  & 90.7  \\
    {SPPnet(ZF net)} & 82.9  & 73.3  & 90.9  & 88.8  & 77.6  & 29.2  & 67.9  & 46.1  & 78.7  & 65.3  & 90.5  & 59.9  & 78.3  & 37.5  & 59.8  & 82.2  \\
    \hline
    \hline
          & cls97 & cls98 & cls99 & cls100 & cls101 & cls102 & cls103 & cls104 & cls105 & cls106 & cls107 & cls108 & cls109 & cls110 & cls111 & cls112 \\
    \hline
    \hline
    {RCNN(CaffeNet)} & 42.6  & 42.0  & 16.8  & 22.4  & 52.4  & 50.4  & 59.8  & 36.2  & 56.8  & 78.3  & 39.6  & 88.7  & 90.8  & 52.7  & 65.3  & 80.5  \\
    {FRCN(VGG16)} & 42.8  & 12.9  & 27.0  & 29.1  & 37.8  & 26.6  & 56.9  & 23.1  & 45.2  & 82.9  & 17.5  & 89.4  & 90.6  & 36.2  & 59.5  & 81.6  \\
    {SPPnet(ZF net)} & 35.7  & 26.8  & 23.0  & 22.9  & 29.0  & 46.3  & 36.3  & 16.0  & 45.0  & 78.8  & 26.3  & 88.9  & 90.8  & 38.2  & 54.8  & 77.4  \\
    \hline
    \hline
          & cls113 & cls114 & cls115 & cls116 & cls117 & cls118 & cls119 & cls120 & cls121 & cls122 & cls123 & cls124 & cls125 & cls126 & cls127 & cls128 \\
    \hline
    \hline
    {RCNN(CaffeNet)} & 79.3  & 89.6  & 57.8  & 79.7  & 90.9  & 77.6  & 75.4  & 89.7  & 45.9  & 69.8  & 89.8  & 68.2  & 55.5  & 61.8  & 79.0  & 52.9  \\
    {FRCN(VGG16)} & 77.5  & 81.8  & 44.1  & 78.1  & 90.9  & 69.3  & 70.2  & 89.2  & 31.3  & 78.9  & 89.9  & 67.5  & 52.5  & 50.6  & 86.1  & 40.3  \\
    {SPPnet(ZF net)} & 67.2  & 78.9  & 34.7  & 75.4  & 90.9  & 54.2  & 60.7  & 82.6  & 23.9  & 72.9  & 88.5  & 55.9  & 31.4  & 44.2  & 80.7  & 38.2  \\
    \hline
    \hline
          & cls129 & cls130 & cls131 & cls132 & cls133 & cls134 & cls135 & cls136 & cls137 & cls138 & cls139 & cls140 & cls141 & cls142 & cls143 & cls144 \\
    \hline
    \hline
    {RCNN(CaffeNet)} & 64.6  & 65.9  & 73.8  & 80.5  & 79.1  & 80.0  & 45.5  & 72.7  & 52.7  & 75.5  & 83.0  & 80.8  & 81.4  & 77.8  & 90.6  & 73.6  \\
    {FRCN(VGG16)} & 62.5  & 53.9  & 67.0  & 79.7  & 78.1  & 79.4  & 32.5  & 63.7  & 62.7  & 71.9  & 77.2  & 72.8  & 89.5  & 66.5  & 90.7  & 64.1  \\
    {SPPnet(ZF net)} & 57.2  & 49.5  & 48.1  & 68.8  & 69.0  & 75.3  & 35.9  & 42.6  & 50.6  & 54.0  & 66.3  & 62.2  & 79.2  & 57.8  & 90.3  & 57.2  \\
    \hline
    \hline
          & cls145 & cls146 & cls147 & cls148 & cls149 & cls150 & cls151 & cls152 & cls153 & cls154 & cls155 & cls156 & cls157 & cls158 & cls159 & cls160 \\
    \hline
    \hline
    {RCNN(CaffeNet)} & 61.2  & 100.0  & 63.3  & 82.9  & 83.3  & 88.3  & 90.9  & 90.5  & 89.8  & 75.3  & 78.1  & 81.4  & 89.6  & 54.0  & 83.6  & 64.8  \\
    {FRCN(VGG16)} & 46.0  & 100.0  & 66.5  & 81.1  & 86.5  & 90.2  & 98.8  & 90.0  & 68.8  & 48.0  & 71.0  & 77.8  & 89.6  & 38.3  & 78.7  & 59.8  \\
    {SPPnet(ZF net)} & 37.3  & 100.0  & 41.2  & 76.4  & 78.9  & 81.5  & 90.4  & 76.0  & 72.5  & 35.0  & 51.9  & 69.6  & 84.1  & 27.2  & 79.3  & 45.3  \\
    \hline
    \hline
    \end{tabular}}%
  \caption{Detailed {\bf average precision (AP) results} of logo detection on the Logos-160 test set. cls1-160 denotes each of the 160 logo classes in our Logo-Net database respectively.}
  \label{table:Detail mAP-160}%
\end{table*}%

\begin{table*}[htbp]
  \centering
    \begin{tabular}{l|*{10}{c}}
    \hline
    Alg(Model) & bnd1  & bnd2  & bnd3  & bnd4  & bnd5  & bnd6  & bnd7  & bnd8  & bnd9  & bnd10 \\
    \hline
    \hline
    {RCNN(CaffeNet)} & 85.0  & 99.1  & 92.2  & 99.1  & 74.0  & 83.0  & 87.5  & 94.3  & 90.1  & 91.9  \\
    {FRCN(VGG16)} & 65.3  & 93.7  & 81.3  & 97.2  & 61.0  & 80.7  & 94.2  & 87.6  & 86.2  & 90.6  \\
    {SPPnet(ZF net)} & 52.0  & 95.9  & 79.9  & 95.9  & 38.3  & 64.3  & 83.9  & 89.1  & 77.1  & 78.1  \\
    \hline
    \hline
          & bnd11 & bnd12 & bnd13 & bnd14 & bnd15 & bnd16 & bnd17 & bnd18 & bnd19 & bnd20 \\
    \hline
    \hline
    {RCNN(CaffeNet)} & 90.8  & 95.1  & 99.4  & 98.3  & 97.2  & 96.0  & 58.6  & 82.3  & 99.3  & 86.8  \\
    {FRCN(VGG16)} & 89.4  & 86.2  & 99.4  & 97.4  & 95.3  & 93.8  & 54.4  & 83.9  & 98.8  & 73.6  \\
    {SPPnet(ZF net)} & 87.8  & 84.7  & 98.4  & 96.3  & 88.0  & 82.9  & 55.3  & 82.7  & 100.0  & 80.9  \\
    \hline
    \hline
          & bnd21 & bnd22 & bnd23 & bnd24 & bnd25 & bnd26 & bnd27 & bnd28 & bnd29 & bnd30 \\
    \hline
    \hline
    {RCNN(CaffeNet)} & 94.9  & 74.0  & 77.4  & 87.7  & 93.6  & 85.4  & 90.4  & 79.8  & 92.5  & 96.8 \\
    {FRCN(VGG16)} & 94.6  & 62.3  & 81.5  & 79.6  & 93.6  & 82.9  & 89.8  & 75.7  & 92.5  & 95.8  \\
    {SPPnet(ZF net)} & 90.9  & 63.7  & 69.0  & 69.1  & 89.1  & 82.6  & 86.3  & 59.8  & 86.0  & 89.3  \\
    \hline
    \hline
          & bnd31 & bnd32 & bnd33 & bnd34 & bnd35 & bnd36 & bnd37 & bnd38 & bnd39 & bnd40 \\
    \hline
    \hline
    {RCNN(CaffeNet)} & 79.1  & 86.1  & 64.8  & 95.6  & 82.4  & 92.7  & 100.0  & 96.8  & 89.2  & 87.4  \\
    {FRCN(VGG16)} & 71.1  & 79.4  & 54.6  & 94.9  & 84.1  & 87.2  & 99.7  & 92.0  & 91.1  & 74.8  \\
    {SPPnet(ZF net)} & 55.3  & 74.9  & 62.1  & 87.8  & 66.5  & 80.7  & 100.0  & 89.6  & 74.6  & 77.0  \\
    \hline
    \hline
          & bnd41 & bnd42 & bnd43 & bnd44 & bnd45 & bnd46 & bnd47 & bnd48 & bnd49 & bnd50 \\
    \hline
    \hline
    {RCNN(CaffeNet)} & 91.5  & 79.9  & 78.6  & 89.9  & 65.3  & 81.7  & 85.2  & 84.3  & 81.5  & 80.3  \\
    {FRCN(VGG16)} & 90.9  & 71.9  & 73.7  & 80.4  & 76.3  & 76.7  & 89.1  & 89.3  & 80.8  & 72.1  \\
    {SPPnet(ZF net)} & 85.4  & 75.9  & 64.1  & 76.6  & 69.9  & 63.7  & 73.8  & 81.2  & 95.3  & 69.1  \\
    \hline
    \hline
          & bnd51 & bnd52 & bnd53 & bnd54 & bnd55 & bnd56 & bnd57 & bnd58 & bnd59 & bnd60 \\
    \hline
    \hline
    {RCNN(CaffeNet)} & 97.4  & 97.8  & 95.5  & 91.4  & 95.7  & 98.5  & 100.0  & 92.9  & 82.8  & 100.0  \\
    {FRCN(VGG16)} & 95.6  & 95.3  & 93.4  & 85.6  & 93.5  & 96.2  & 98.8  & 90.8  & 81.9  & 100.0  \\
    {SPPnet(ZF net)} & 94.4  & 97.2  & 89.8  & 80.4  & 91.1  & 92.3  & 99.3  & 83.5  & 70.5  & 100.0  \\
    \hline
    \hline
          & bnd61 & bnd62 & bnd63 & bnd64 & bnd65 & bnd66 & bnd67 & bnd68 & bnd69 & bnd70 \\
    \hline
    \hline
    {RCNN(CaffeNet)} & 90.8  & 86.5  & 85.3  & 76.2  & 72.5  & 98.3  & 98.2  & 94.0  & 97.5  & 94.4  \\
    {FRCN(VGG16)} & 86.0  & 92.4  & 83.8  & 70.7  & 68.0  & 99.3  & 96.0  & 94.2  & 94.1  & 87.2  \\
    {SPPnet(ZF net)} & 80.8  & 67.2  & 69.8  & 69.3  & 62.9  & 96.5  & 96.3  & 84.2  & 93.5  & 89.4  \\
    \hline
    \hline
          & bnd71 & bnd72 & bnd73 & bnd74 & bnd75 & bnd76 & bnd77 & bnd78 & bnd79 & bnd80 \\
    \hline
    \hline
    {RCNN(CaffeNet)} & 88.8  & 99.0  & 96.9  & 92.6  & 95.7  & 90.5  & 78.6  & 92.8  & 86.1  & 91.7  \\
    {FRCN(VGG16)} & 76.9  & 98.7  & 92.0  & 85.8  & 93.9  & 84.1  & 56.3  & 93.6  & 88.0  & 82.7  \\
    {SPPnet(ZF net)} & 70.6  & 98.7  & 95.9  & 92.2  & 89.9  & 87.2  & 55.5  & 90.2  & 81.7  & 75.0  \\
    \hline
    \hline
          & bnd81 & bnd82 & bnd83 & bnd84 & bnd85 & bnd86 & bnd87 & bnd88 & bnd89 & bnd90 \\
    \hline
    \hline
    {RCNN(CaffeNet)} & 95.4  & 94.9  & 82.1  & 88.7  & 87.8  & 93.9  & 99.5  & 100.0  & 97.8  & 100.0  \\
    {FRCN(VGG16)} & 94.6  & 95.4  & 72.9  & 92.1  & 66.3  & 85.5  & 98.4  & 100.0  & 94.7  & 100.0  \\
    {SPPnet(ZF net)} & 87.6  & 92.4  & 50.5  & 75.2  & 75.6  & 78.4  & 97.2  & 99.5  & 89.6  & 100.0  \\
    \hline
    \hline
          & bnd91 & bnd92 & bnd93 & bnd94 & bnd95 & bnd96 & bnd97 & bnd98 & bnd99 & bnd100\\
    \hline
    \hline
    {RCNN(CaffeNet)} & 80.5  & 98.4  & 98.0  & 98.8  & 99.7  & 95.6  & 87.3  & 98.6  & 77.3  & 75.2 \\
    {FRCN(VGG16)} & 79.2  & 93.5  & 96.1  & 99.1  & 99.7  & 56.0  & 66.9  & 97.5  & 65.7  & 55.2  \\
    {SPPnet(ZF net)} & 69.3  & 88.2  & 93.8  & 98.5  & 100.0  & 90.2  & 70.3  & 95.4  & 60.7  & 73.4  \\
    \hline
    \hline
    \end{tabular}%
    \caption{Detailed {\bf brand recognition accuracy} results of 100-brand recognition on the Logos-160 test set. Bnd1-100 denotes each of the 100 brand categories in our Logo-Net database respectively.}
  \label{table:Detail Accuracy-160}%
\end{table*}%

\begin{table*}[htbp]
  \centering
  \small
  \renewcommand{\arraystretch}{2}
    \begin{tabular}{|l|r|l|r|l|r|l|r|}
    \hline
    brand name & \#images & brand name & \#images & brand name & \#images & brand name & \#images \\
    \hline
    \hline
    3T    & 289   & coco  & 547   & kingston & 1100  & otterbox & 915 \\
    \hline
    Girdear & 531   & converse & 2259  & kisscat & 534   & patek philippe & 1059 \\
    \hline
    IWC   & 580   & d\&g  & 892   & l.f.c & 1087  & paul frank & 650 \\
    \hline
    Manchester\_United & 534   & dior  & 513   & lacoste & 921   & pinarello & 1037 \\
    \hline
    Piaget & 501   & dunhill & 515   & lamborghini & 1176  & prada & 500 \\
    \hline
    UGG   & 501   & edwin & 583   & lee   & 564   & puma  & 982 \\
    \hline
    YSL   & 1016  & element case & 566   & lego  & 976   & ralph lauren & 1014 \\
    \hline
    abercrombie\&fitch & 681   & esprit & 489   & levi's & 1002  & rapha & 455 \\
    \hline
    adidas & 977   & evisu & 493   & loewe & 984   & rayban & 653 \\
    \hline
    air Jordan & 534   & fendi & 977   & longines & 538   & rolex & 534 \\
    \hline
    armani & 1755  & ferragamo & 546   & louis vuitton & 789   & samsonite & 684 \\
    \hline
    arsenal & 766   & fiveplus & 660   & mcm   & 687   & sandisk & 1024 \\
    \hline
    asics & 528   & focus t25 & 209   & michael Kors & 933   & septwolves & 622 \\
    \hline
    beats & 577   & fsa   & 355   & miumiu & 484   & spy   & 684 \\
    \hline
    belle & 527   & g-shock & 529   & mulberry & 806   & st\&sat & 530 \\
    \hline
    blackberry & 535   & gap   & 548   & new balance & 897   & stuart weitzman & 523 \\
    \hline
    bottega veneta & 600   & goyard & 1000  & new era & 1124  & tata  & 536 \\
    \hline
    calvin klein & 863   & gucci & 717   & nike  & 893   & teemix & 509 \\
    \hline
    camel & 697   & guess & 527   & nikon & 540   & the\_north\_face & 542 \\
    \hline
    cartier & 799   & harley davidson & 653   & nintendo & 927   & tissot & 568 \\
    \hline
    casio & 585   & hermes & 1476  & nissan & 779   & tommy hilfiger & 1047 \\
    \hline
    celine & 500   & honda & 1079  & old navy & 653   & tudor & 539 \\
    \hline
    chanel & 816   & iphone & 836   & omega & 646   & vans  & 981 \\
    \hline
    coach & 898   & joy\&peace & 496   & only  & 515   & versace & 1007 \\
    \hline
    coca cola & 629   & kate\_spade & 552   & oakley & 556   & zenith & 477 \\
    \hline
    \end{tabular}%
    \renewcommand{\arraystretch}{1.0}
    \caption{Statistics of product images for the 100-brand categories in the Logos-160 dataset}%
  \label{table:dataset-160-brand}%
\end{table*}%

\begin{figure*}[htb]
\vspace{-1in}
\hspace{-0.5in}\scalebox{1.00}{
{\hspace{0in}\includegraphics[width=1.2\linewidth]{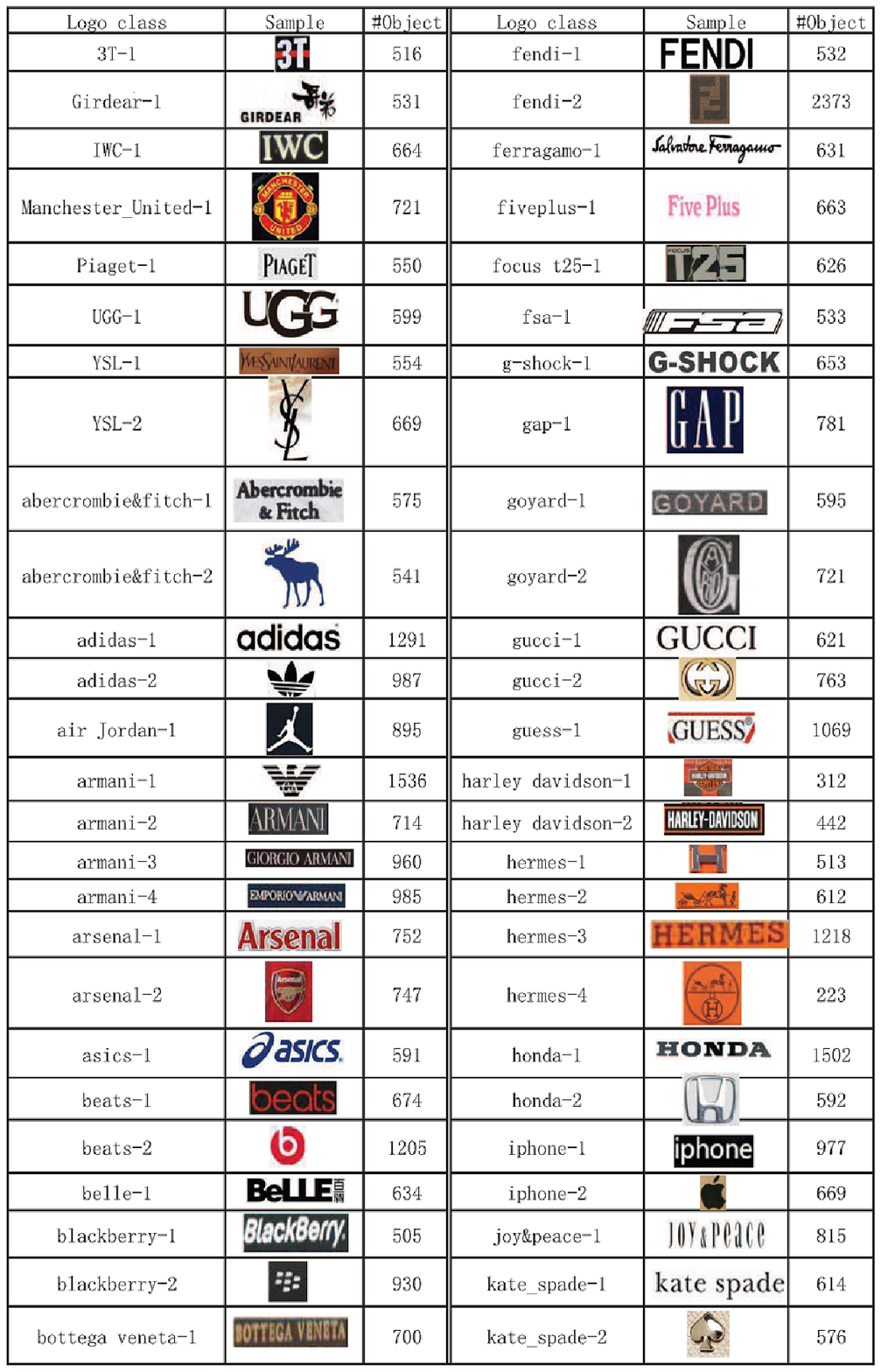}\vspace{-0.2in}}
}
\vspace{-0.1in}
\end{figure*}

\begin{figure*}[htb]
\vspace{-1in}
\hspace{-0.5in}\scalebox{1.0}{
\includegraphics[width=1.2\linewidth]{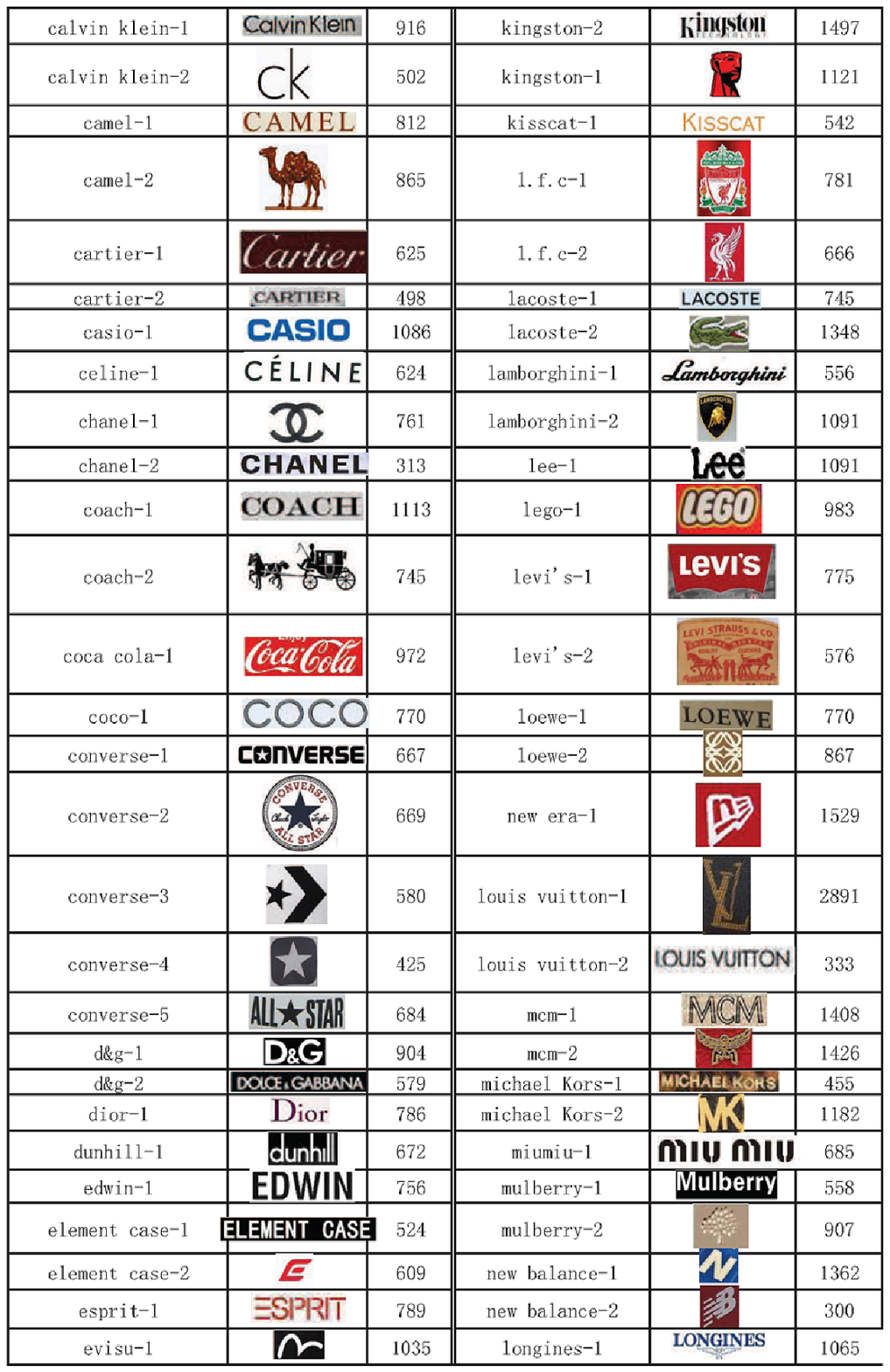}\vspace{-0.2in}
}
\vspace{-0.1in}
\end{figure*}

\end{document}